%% file: main.tex
\documentclass{article}

\usepackage[utf8]{inputenc} 
\usepackage{url}            
\usepackage{booktabs}       
\usepackage{amsfonts}       
\usepackage{nicefrac}       
\usepackage{microtype}      
\usepackage{lipsum}
\usepackage{enumitem}
\usepackage{url}            
\usepackage{graphicx}
\usepackage{arxiv}
\usepackage[authoryear]{natbib}
\usepackage{subcaption}
\usepackage[skins]{tcolorbox}
\usepackage{hyperref}
\usepackage[T1]{fontenc}
\usepackage{epigraph}
\usepackage[export]{adjustbox}
\usepackage{amsmath}
\usepackage{amssymb,mathtools}
\usepackage{multirow}
\usepackage{bbm}

\hypersetup{
    colorlinks=false,
    linkcolor=blue,
    filecolor=magenta,      
    urlcolor=blue,
    citecolor=blue
}

\newcommand{\pctrange}[2]{$#1\%\pm#2$}

\makeatletter
\newcommand{\printfnsymbol}[1]{%
  \textsuperscript{\@fnsymbol{#1}}%
}
\makeatother

\title{
Randomness in Neural Network Training: Characterizing the Impact of Tooling}

\author{%
  Donglin Zhuang \\
  School of Computer Science\\
  The University of Sydney\\
  \texttt{dzhu9887@sydney.edu.au} \\
   \And
   Xingyao Zhang \\
   Department of Computer Science \\
   University of Washington \\
   \texttt{xingyaoz@cs.washington.edu} \\
   \AND
   Shuaiwen Leon Song \\
   School of Computer Science \\
   The University of Sydney \\
   \texttt{shuaiwen.song@sydney.edu.au} \\
   \And
   Sara Hooker \\
   Google Research, Brain \\
   \texttt{shooker@google.com} \\
}

\begin{document}

\maketitle

\begin{abstract}
The quest for determinism in machine learning has disproportionately focused on characterizing the impact of noise introduced by algorithmic design choices. In this work, we address a less well understood and studied question: how does our choice of tooling introduce randomness to deep neural network training. We conduct large scale experiments across different types of hardware, accelerators, state of art networks, and open-source datasets, to characterize how tooling choices contribute to the level of non-determinism in a system, the impact of said non-determinism, and the cost of eliminating different sources of noise.

Our findings are surprising, and suggest that the impact of non-determinism in nuanced. While top-line metrics such as top-1 accuracy are not noticeably impacted, model performance on certain parts of the data distribution is far more sensitive to the introduction of randomness. Our results suggest that deterministic tooling is critical for AI safety. However, we also find that the cost of ensuring determinism varies dramatically between neural network architectures and hardware types, e.g., with overhead up to \textit{746\%, 241\%, and 196\%} on a spectrum of widely used GPU accelerator architectures, relative to non-deterministic training.  The source code used in this paper is available at https://github.com/usyd-fsalab/NeuralNetworkRandomness.
\end{abstract}











\section{Introduction}

In the pursuit of scientific progress, a key desiderata is to eliminate noise from a system. As scientists, we typically regard noise as all the random variations \emph{independent} of the signal we are trying to measure. In the field of machine learning, the urgency to remove noise from training is often motivated by 1) concerns around replicability of experiment results, 2) having full experimental control and/or 3) the need to precisely audit AI behavior in safety-critical domains where human welfare may be harmed.

Recent work has disproportionately focused on the impact of algorithm design choices on model replicability \citep{impactofnondeterminismrl,madhyastha-jain-2019-model,summers2021nondeterminism,Snapp2021,shamir2020smooth,lucic2018gans,Henderson2017}. Less well explored or understood is how our choice of tooling impacts the level of noise in a machine learning system. While some recent work has evaluated the role of software dependencies \citep{Pham2020,Hong2013}, this has been evaluated in the context of a single machine. In parallel, the quest for determinism has spurred the design of hardware that is inherently deterministic  \cite{gpudet,dab,jouppi2017indatacenter} and software patches that ensure determinism in popular deep learning libraries such as Tensorflow \citep{tensorflow}, Jax \citep{jax2018github}, Pytorch \citep{pytorch}, and cuDNN \citep{cudnn}.

\begin{figure*}
	\centering 
	\vskip 0.15in
    \begin{small}
    \begin{sc}
    \begin{subfigure}{0.495\linewidth}
        \begin{subfigure}{0.33\linewidth}
		    \centering \includegraphics[width=0.99\linewidth]{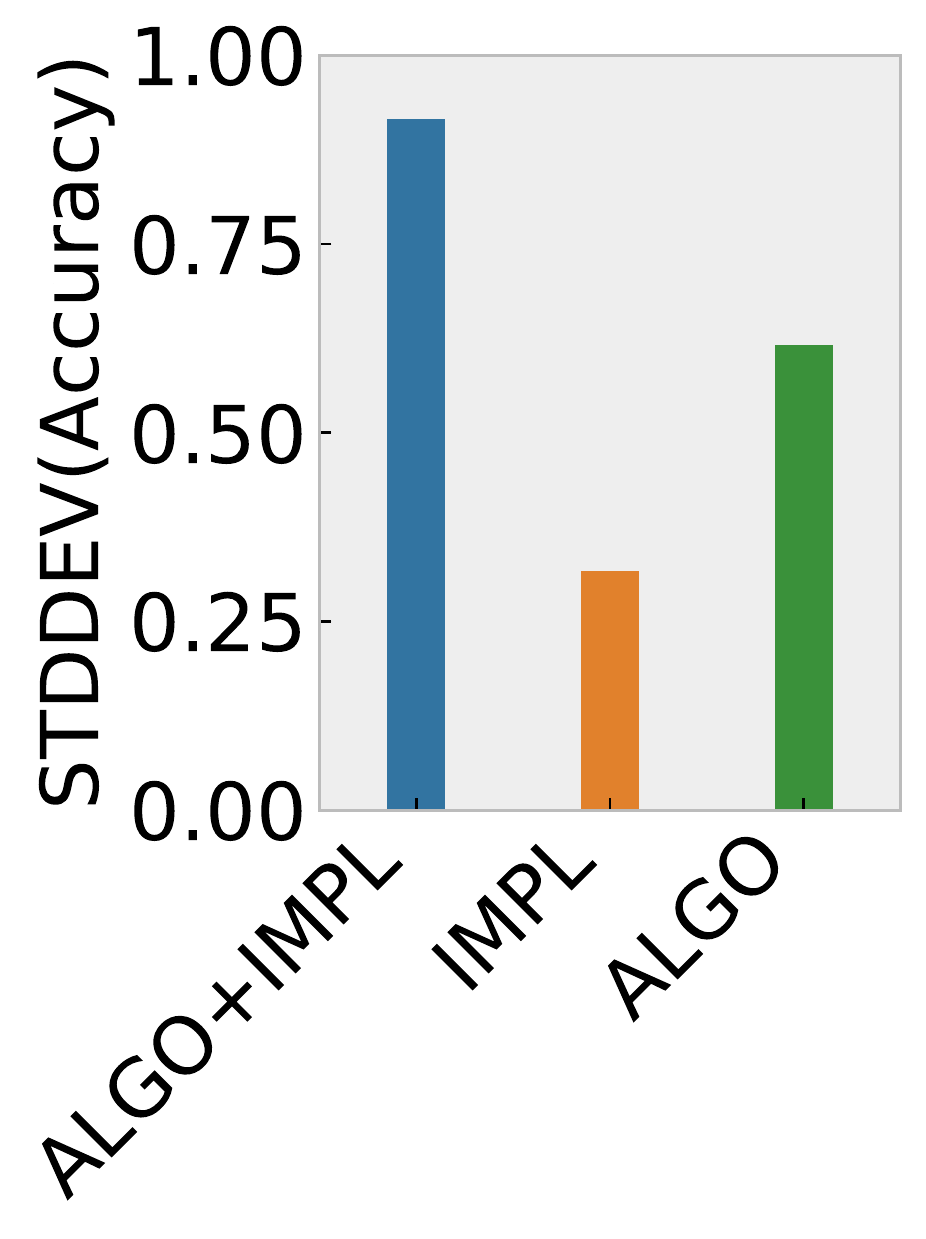}
        \end{subfigure}	
        \hspace{-0.5em}
        \begin{subfigure}{0.33\linewidth}
        	\centering \includegraphics[width=0.99\linewidth]{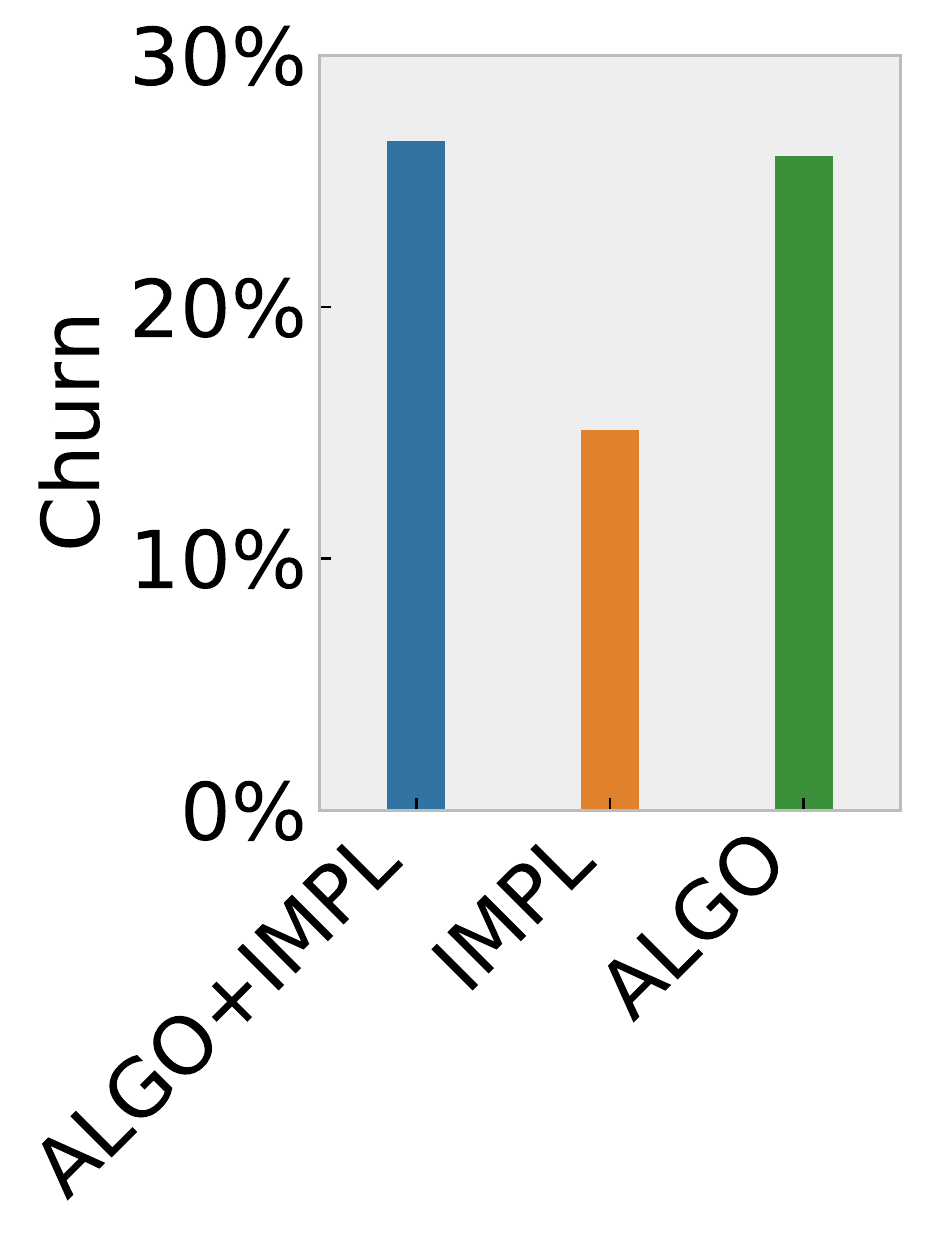}
        \end{subfigure}
        \hspace{-0.5em}
        \begin{subfigure}{0.33\linewidth}
        	\centering \includegraphics[width=0.99\linewidth]{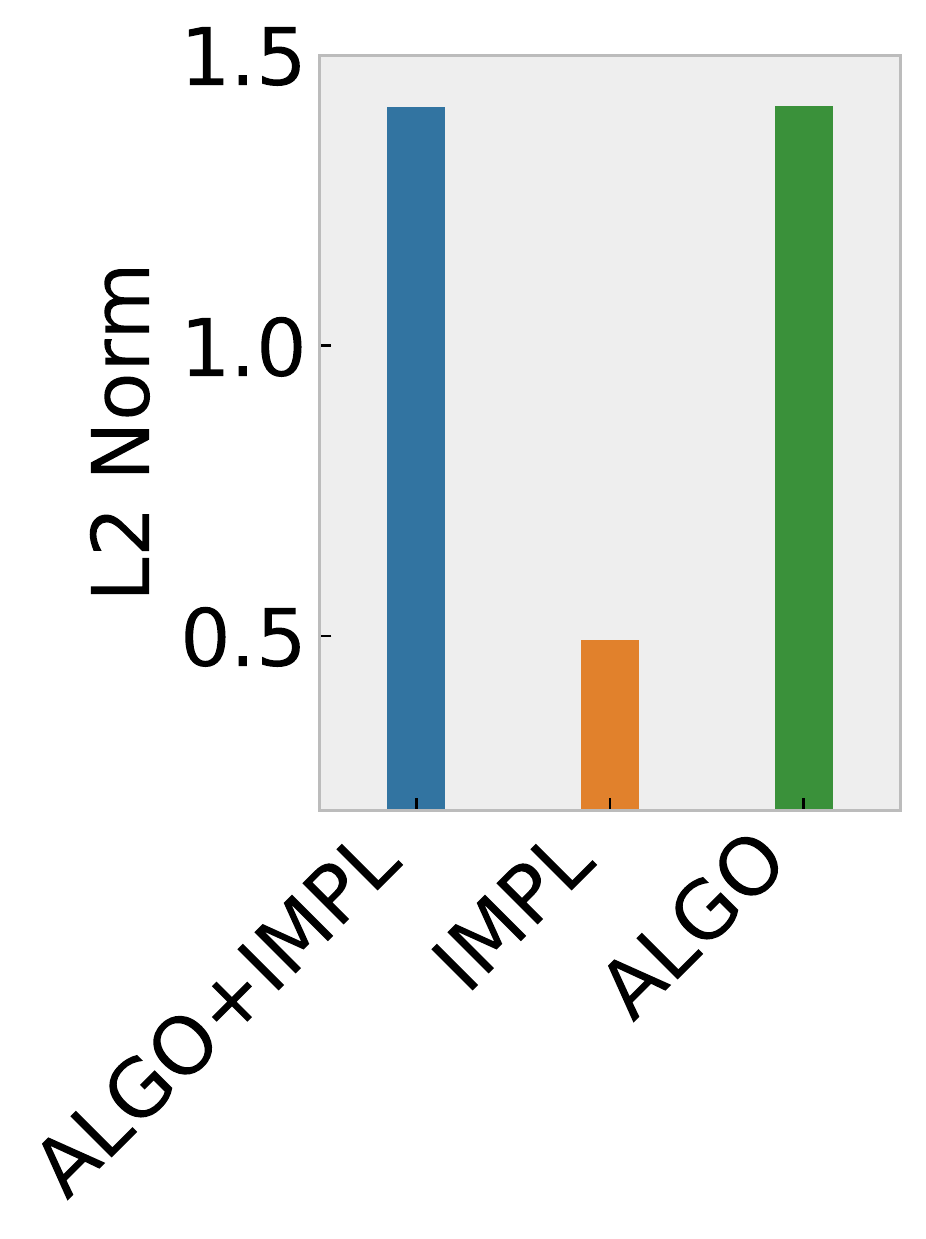}
        \end{subfigure}
        \caption{\textbf{Small CNN CIFAR-10}}
    \end{subfigure}
    \hfill
    \begin{subfigure}{0.495\linewidth}
        \begin{subfigure}{0.33\linewidth}
    		\centering \includegraphics[width=0.99\linewidth]{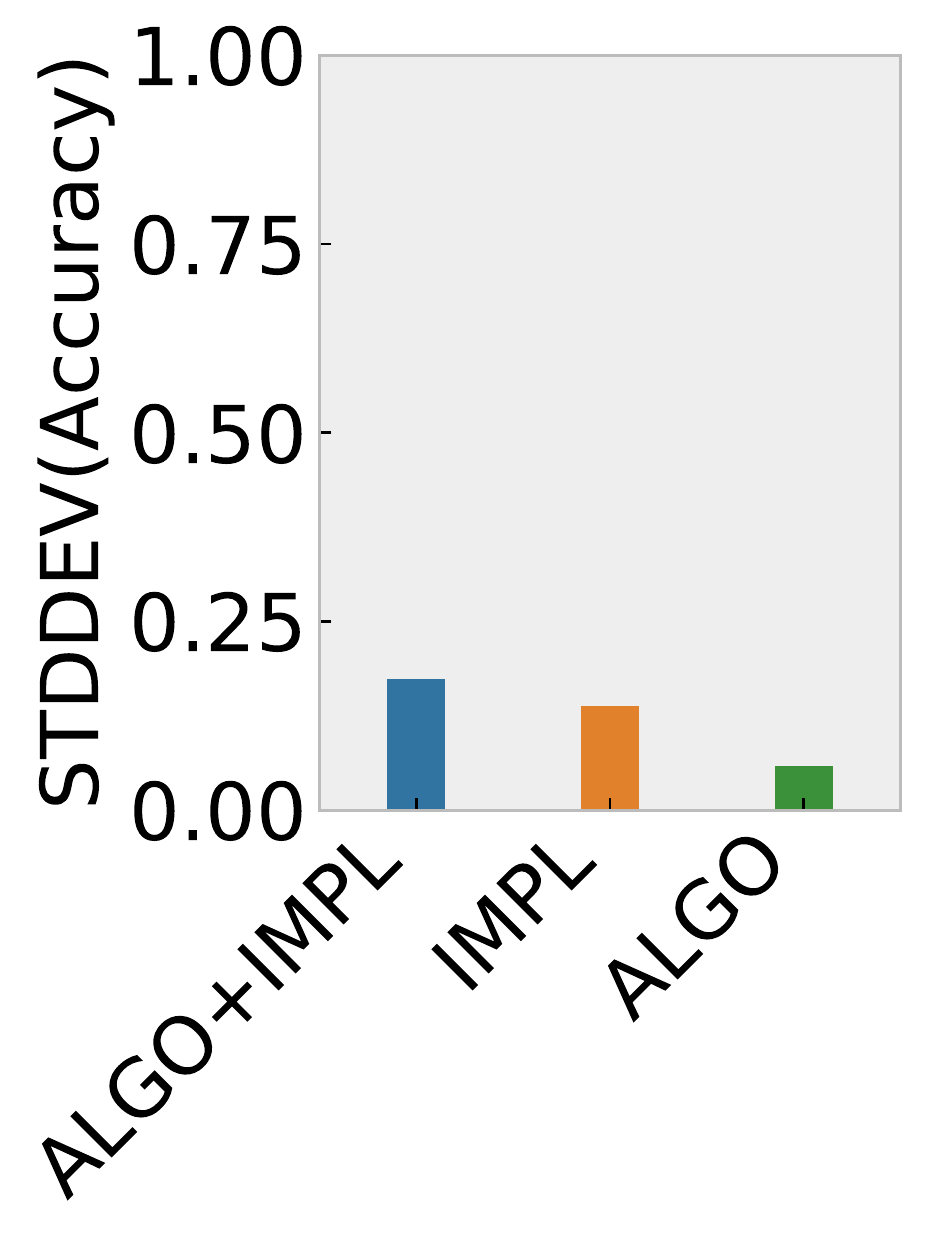}
        \end{subfigure}	
        \hspace{-0.5em}
        \begin{subfigure}{0.33\linewidth}
        	\centering	\includegraphics[width=0.99\linewidth]{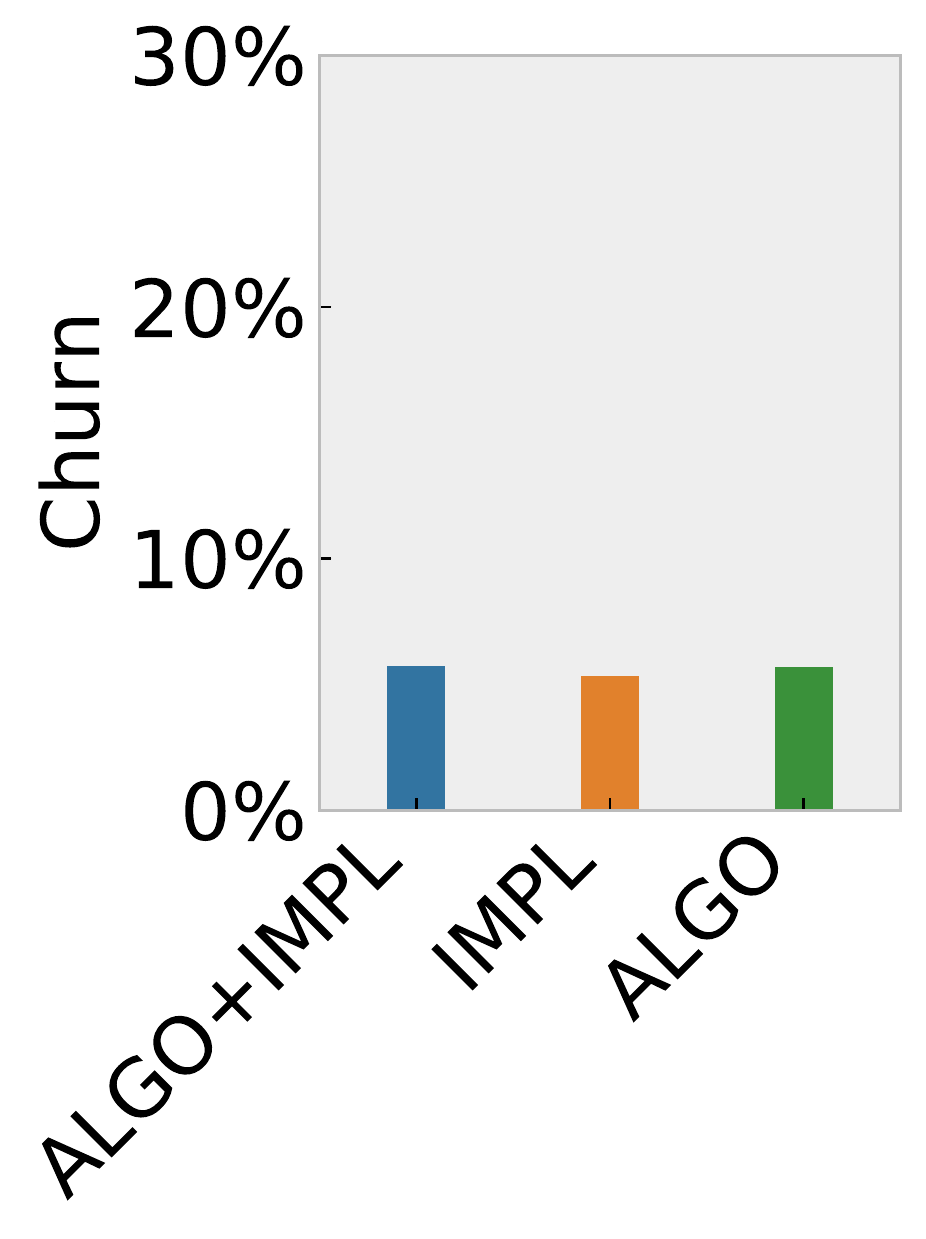}
        \end{subfigure}
        \hspace{-0.5em}
        \begin{subfigure}{0.33\linewidth}
        	\centering \includegraphics[width=0.99\linewidth]{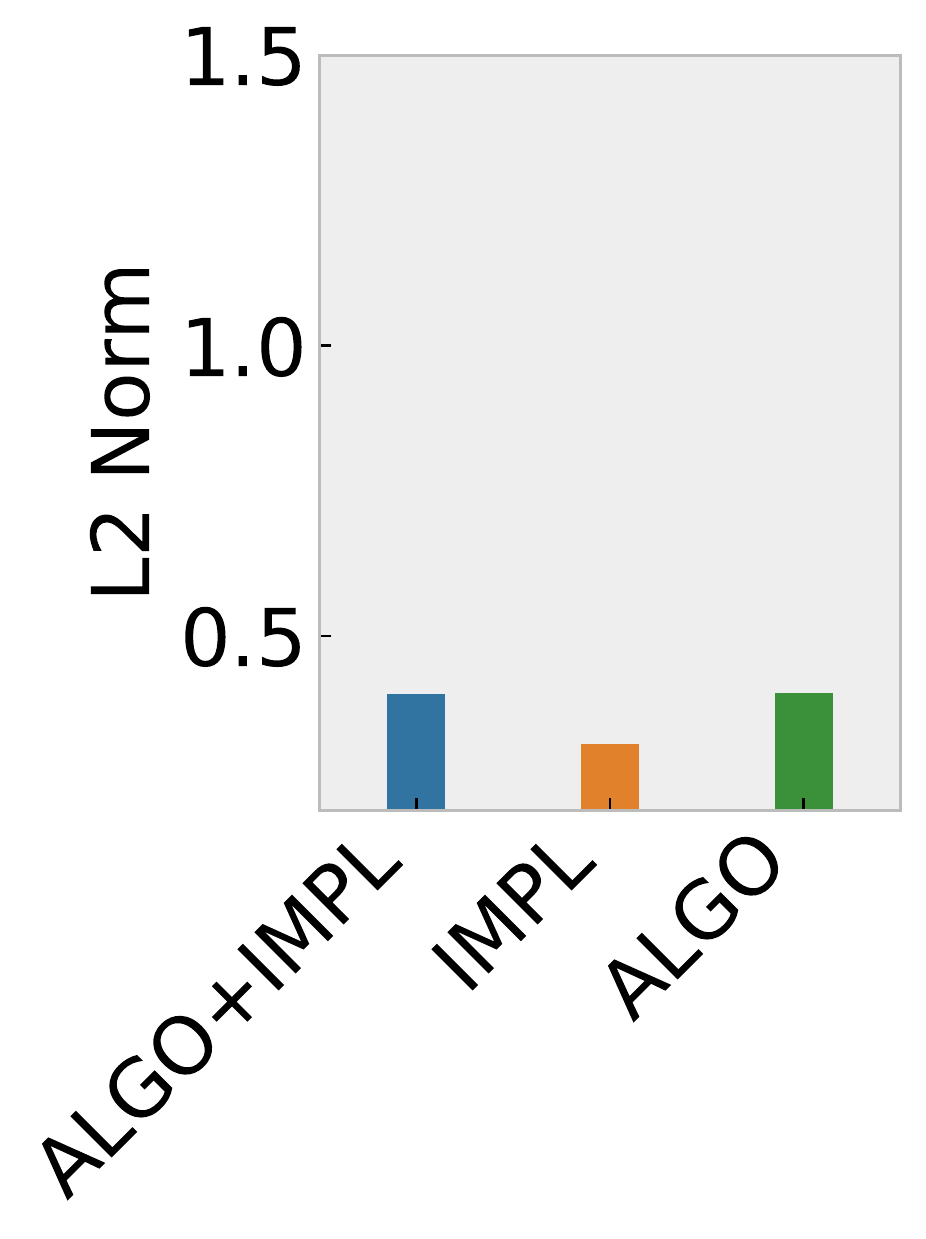}
        \end{subfigure}
               \caption{\textbf{ResNet-18 CIFAR-10}}
    \end{subfigure}
    \\
    \vskip 0.15in
    \begin{subfigure}{0.495\linewidth}
         \begin{subfigure}{0.33\linewidth}
    		\centering \includegraphics[width=0.99\linewidth]{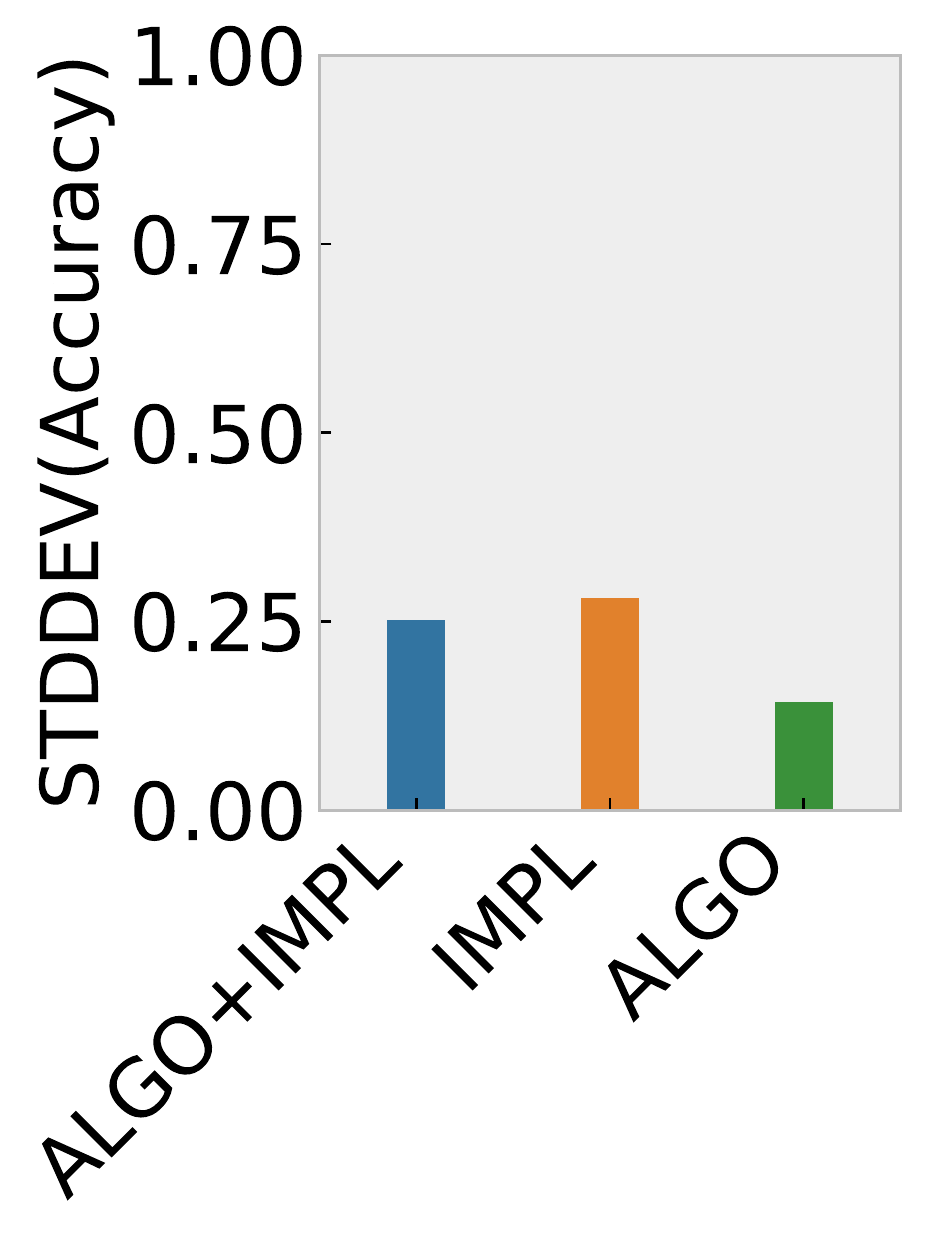}
        \end{subfigure}	
        \hspace{-0.5em}
        \begin{subfigure}{0.33\linewidth}
        	\centering	\includegraphics[width=0.99\linewidth]{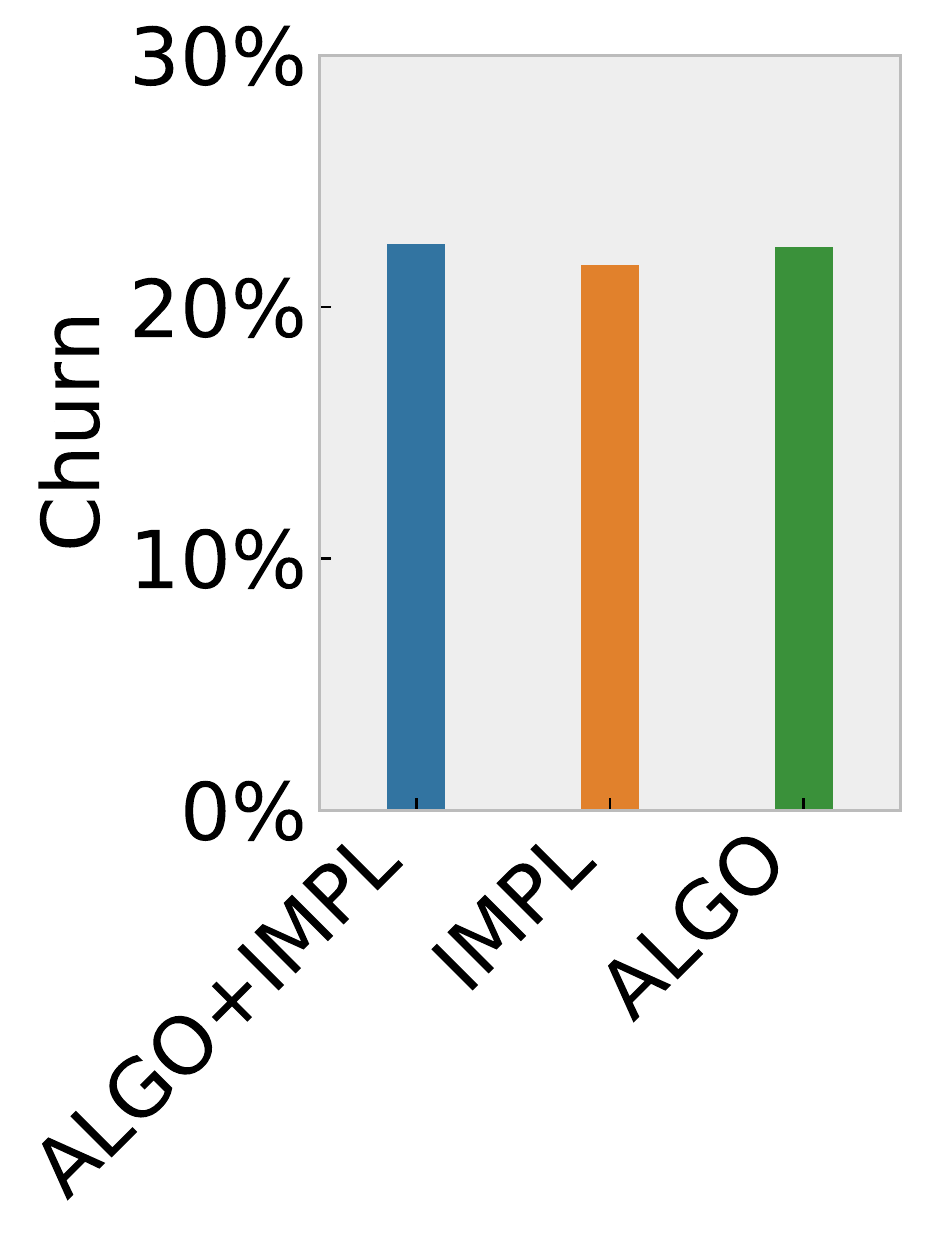}
        \end{subfigure}
        \hspace{-0.5em}
        \begin{subfigure}{0.33\linewidth}
        	\centering \includegraphics[width=0.99\linewidth]{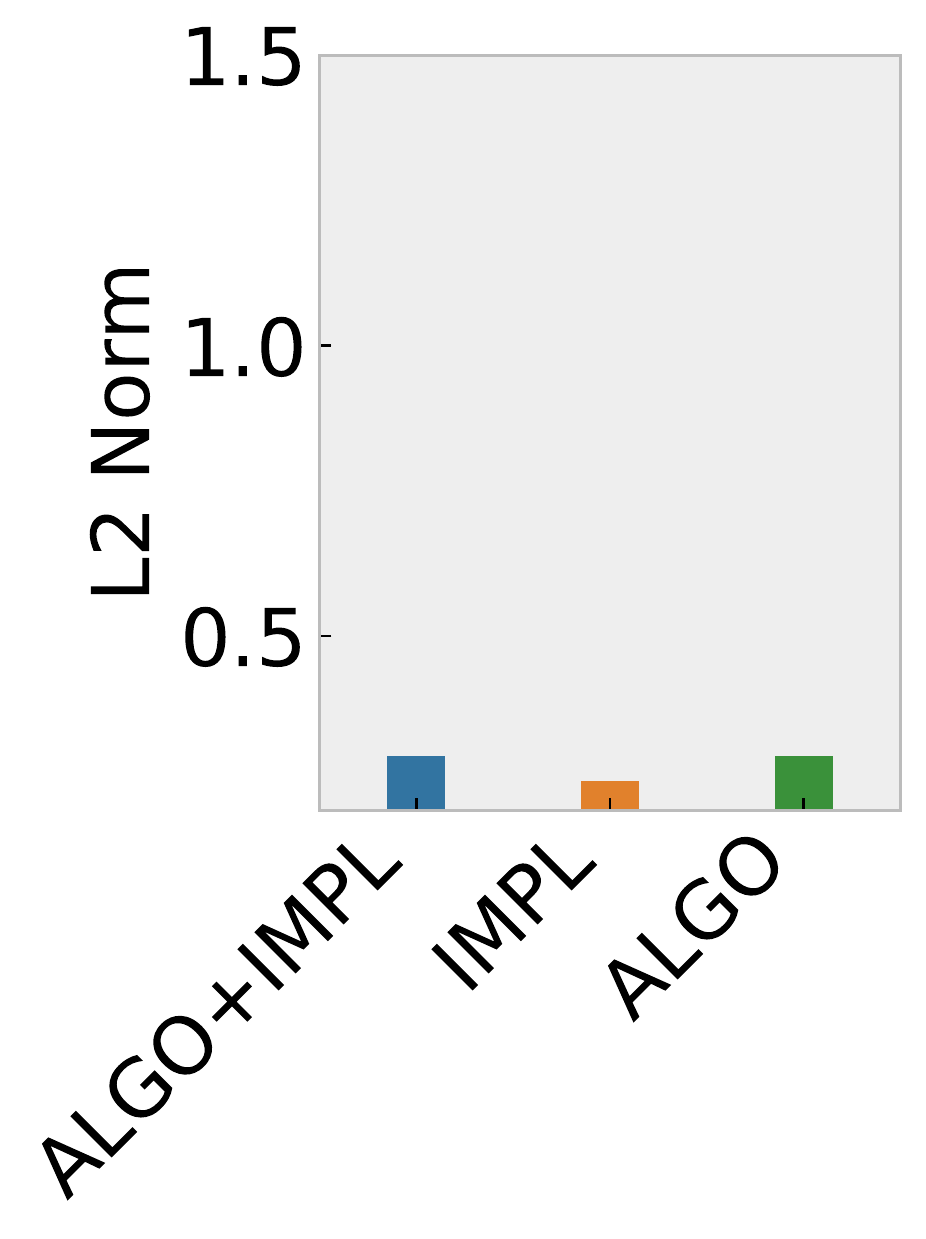}
        \end{subfigure}
        \caption{\textbf{ResNet-18 CIFAR-100}}
    \end{subfigure}
    \hfill
    \begin{subfigure}{0.495\linewidth}
        \begin{subfigure}{0.33\linewidth}
    		\centering \includegraphics[width=0.99\linewidth]{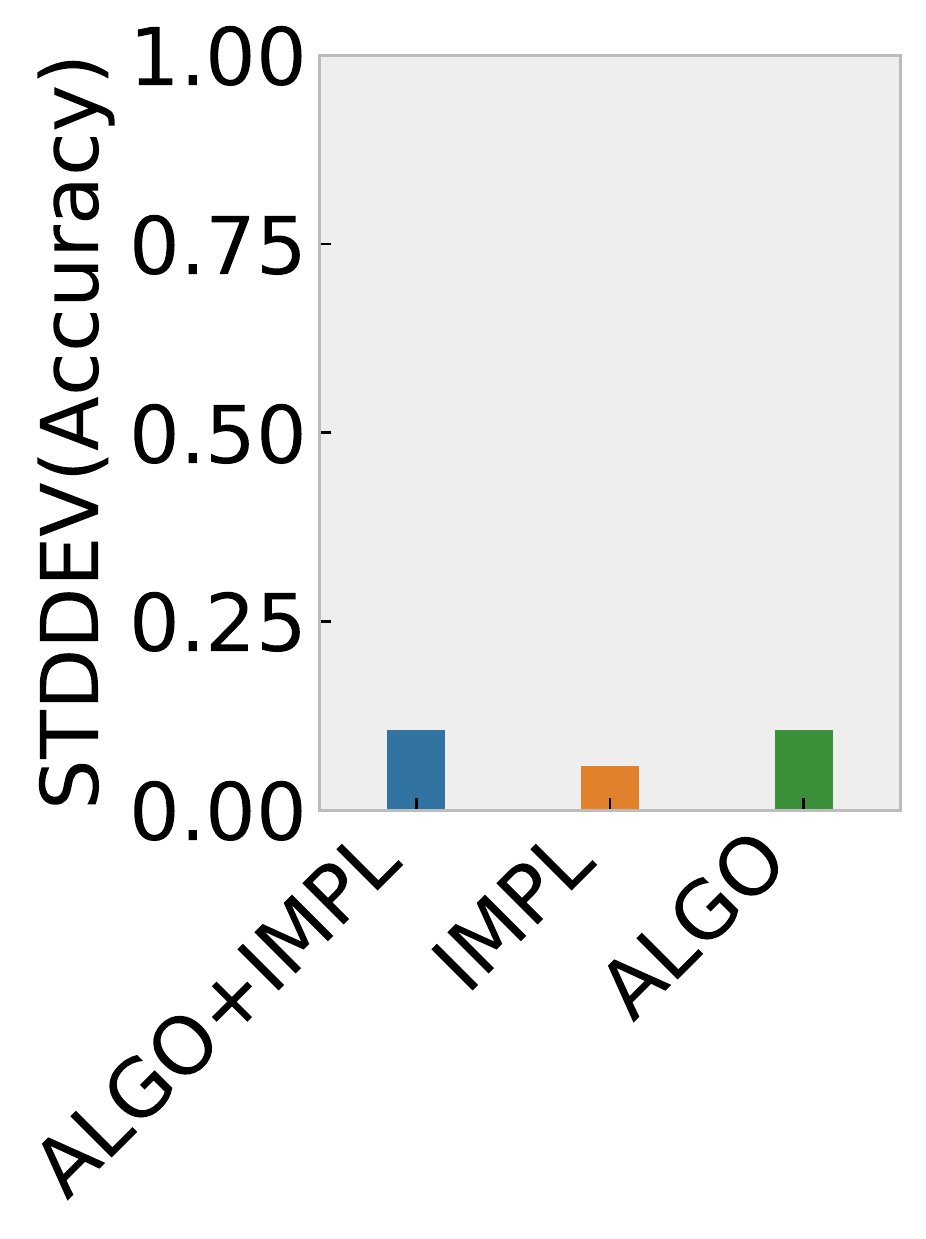}
        \end{subfigure}	
        \hspace{-0.5em}
        \begin{subfigure}{0.33\linewidth}
        	\centering	\includegraphics[width=0.99\linewidth]{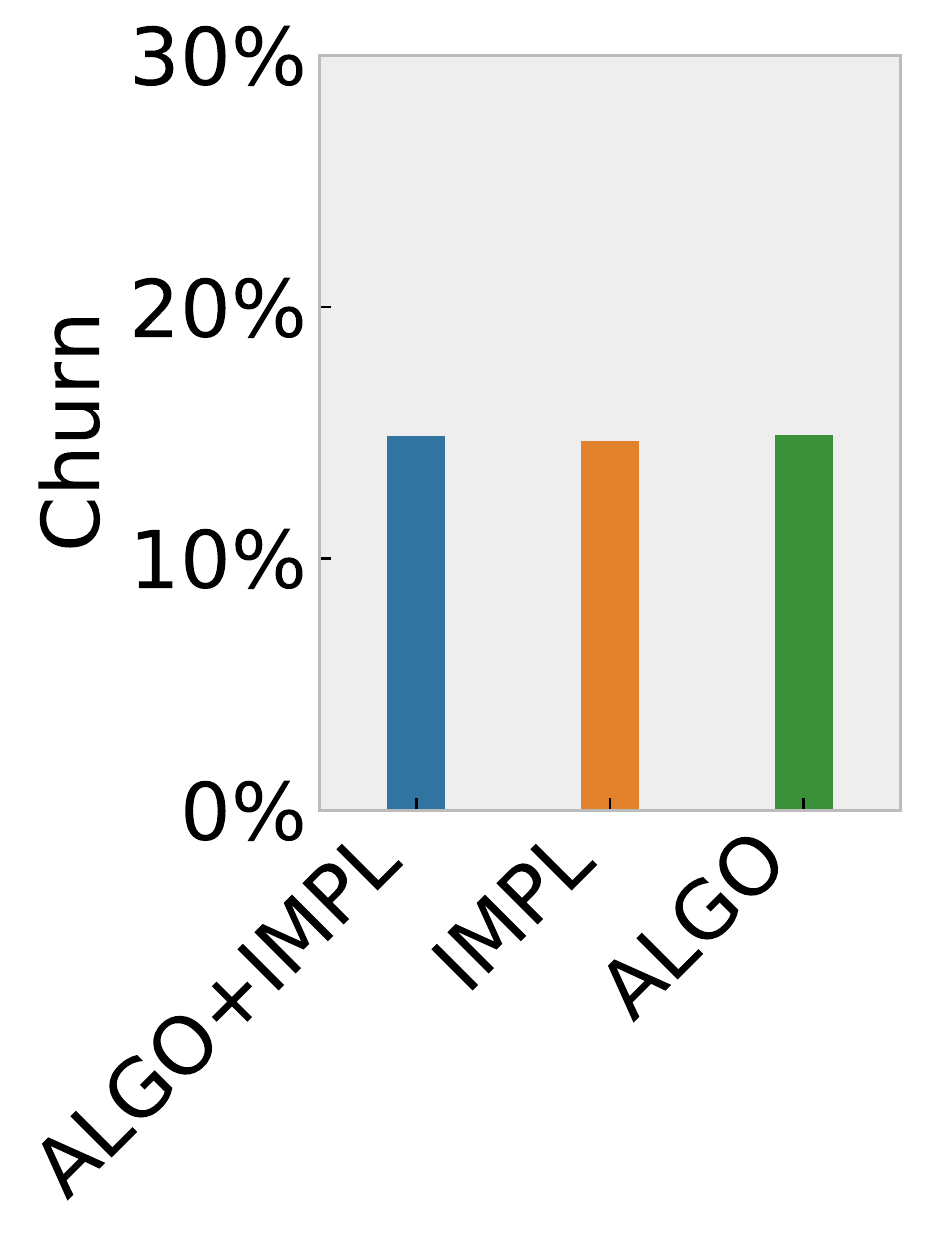}
        \end{subfigure}
        \hspace{-0.5em}
        \begin{subfigure}{0.33\linewidth}
        	\centering \includegraphics[width=0.99\linewidth]{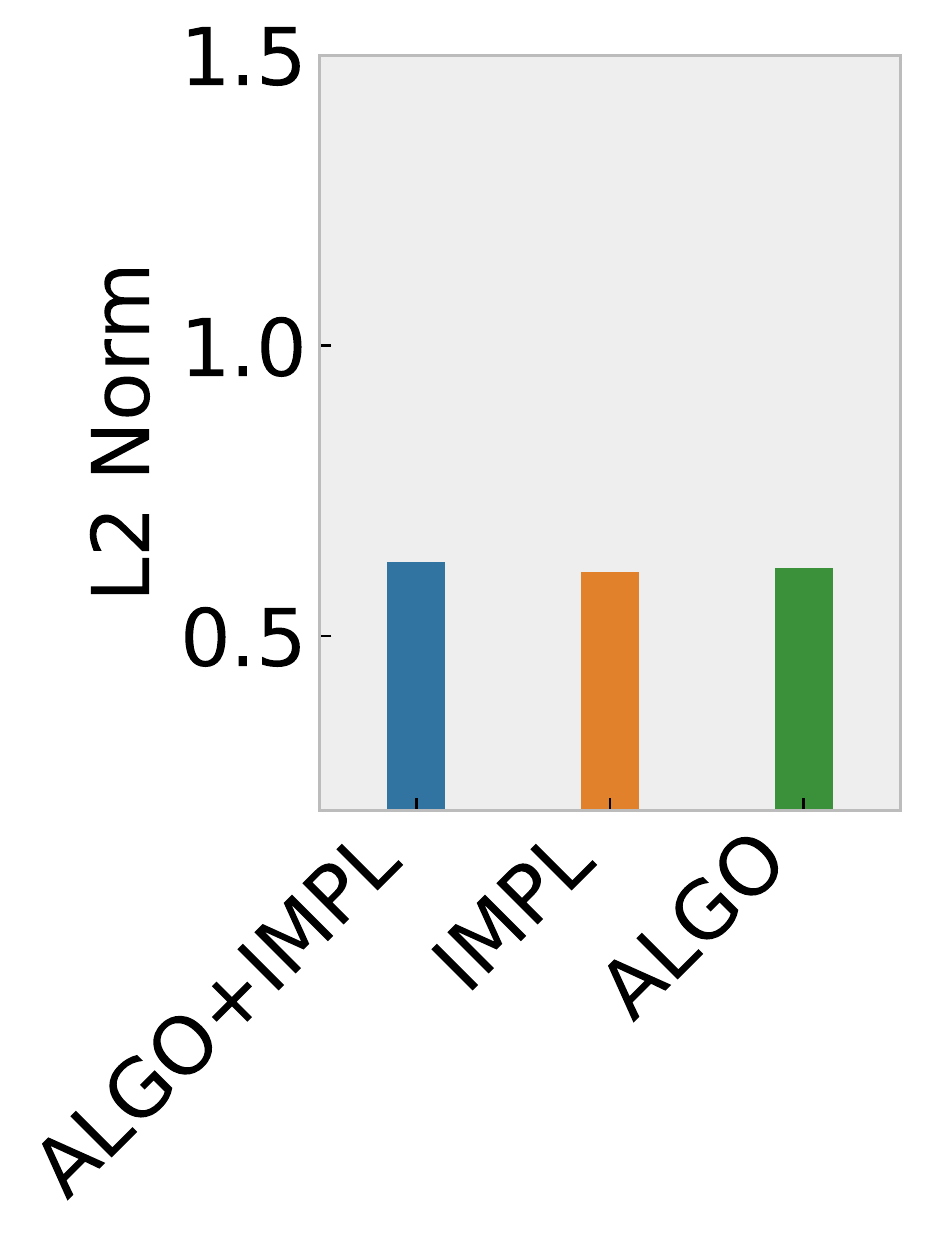}
        \end{subfigure}
        \caption{\textbf{ResNet-50 ImageNet}}
    \end{subfigure}
    
	\end{sc}
	\end{small}
	\caption{
	 Comparison of different source of noise on standard deviation of accuracy, predictive churn and L2 distance between trained weights (on V100 GPU). \textit{Implementation noise} (IMPL) introduces less uncertainty than \textit{algorithmic noise} (ALGO) in terms of Churn and L2 distance, but each is a significant source of uncertainty}
	\label{fig:types_noise_v100}
		\vskip -0.15in
\end{figure*}

In our rush to eliminate noise from ML systems, we seem to have skipped a crucial step -- characterizing the origins of the problem and the cost of controlling noise in the system. Understanding the sources of noise in ML systems and the downstream impact is critical in order to weigh the benefits of controlling noise at different levels of the technology stack. How does the choice of hardware, software and algorithm individually contribute to the overall system-level noise? Here, we seek to identify individual sources of randomness at different levels of the technology stack. We separately isolate and evaluate the contribution of both \textit{algorithmic choices} (i.e. random initialization, data shuffling, random layers and stochastic data augmentation), and \textit{implementation choices} which is the combination of hardware and software used to train the model. Our work is the first to our knowledge to evaluate the impact of different widely used hardware types, and also quantify differences in the cost of controlling noise across hardware. 

Our results are surprising, and suggest that a more nuanced understanding of noise can also inform our understanding of how our tooling impacts generalization. We find that both algorithmic and hardware factors exert minimal difference in top-line metrics. However, we observe a far more pronounced impact on the level of predictive divergence between different model runs, the standard deviation of per-class metrics and sub-group performance. Here, we find that the presence of noise can amplify uncertainty disproportionately on certain subsets of the dataset. While models maintain similar top-line metrics, randomness present during training often causes unacceptable differences in performance on subsets of the population. Notably, we find that non-determinism at all levels of the technology stack can amplify model bias by disproportionately increasing variance in performance on underrepresented sensitive sub-groups.

Our results suggest that deterministic tooling is critical for ensuring AI safety in sensitive domains such as credit scoring, health care diagnostics \citep{2019Hongtao, Gruetzemacher20183DDL, 2019badgeley,2019oakden} and autonomous driving \citep{2017Telsa}.  However, our work also establishes that the cost of fully ensuring determinism is large and \emph{highly} variably due to the sensitivty to model design and underlying hardware. Controlling implementation noise comes with non-negligible training speed overhead for which researchers should weigh the price and benefit based on their tolerance of uncertainty and the sensitivity of the task.
%


Our core contributions can be enumerated as follows:
\begin{enumerate}
\itemsep0em
    \item We establish a rigorous framework for evaluating the impact of tooling on different measures of model stability. We establish consistent results across an extensive experimental set-up, conducting large-scale experiments across different hardware, accelerators, widely used training architectures and datasets (Section~\ref{sec:result:topline-metric}).
    \item  \textit{Non-determinism must be controlled at all levels of the technical stack or is not worth controlling at all.} Even if algorithmic factors are controlled, the noise from tooling alone is substantial. This suggests that removing a single source of noise cannot effectively reduce the level of uncertainty of trained models (Section~\ref{sec:result:sub-group-performance}). The overall level of system noise is highly dependent on model design, with choices such as the presence of batch-normalization \citep{Ioffe2015} driving differences in model stability.  
    \item \textit{Non-determinism has a pronounced impact on sub-aggregate measures of model stability.} While we observe minimal impact on top-line metrics, we find that model performance on certain sub-sets of the distribution is far more sensitive, with underrepresented attributes disproportionately impacted by the introduction of stochasticity (Section~\ref{sec:result:sub-group-performance}).
    \item \textit{Large variance in overhead introduced by deterministic training.} Controlling for implementation noise poses significant overhead to model training procedures -- with overhead up to \textit{746\%, 241\%, and 196\%} on a spectrum of widely used GPU accelerator architectures, relative to non-deterministic training (Section~\ref{sec:result:overhead}). 
\end{enumerate}

\section{Methodology}\label{sec:methodology}

\begin{table*}[t]
\begin{center}
\begin{small}
\begin{tabular}{p{0.20\linewidth}|p{0.70\linewidth}}
\multicolumn{2}{c}{\textbf{Algorithmic Sources of Randomness}}    \\
\toprule
\textbf{Source} & \textbf{Method} \\
Random Initialization &  weights initialized by sampling random distribution \citep{pmlr-v9-glorot10a,resnet} \\
\midrule
Data augmentation & stochastic transformations to the input data \citep{kukacka2017regularization,Hern_ndez_Garc_a_2018,dwibedi2017cut,zhong2017random}  \\
\midrule
Data shuffling & inputs shuffled randomly and batched during training  \citep{smith2018dont} \\
\midrule
Stochastic Layers & e.g. dropout \citep{Srivastava2014,hinton2012improving,pmlr-v28-wan13}, noisy activations \citep{10.5555/3104322.3104425} \\
\midrule
\end{tabular}
\end{small}
 \end{center}
 \caption{Overview of different sources of algorithm (\texttt{ALGO}) noise.}
 \label{table:source_randomness} \end{table*}

We consider a supervised learning setting, 

\begin{equation}
    \mathcal{D} \big\{ (x_1 , y_1), \ldots , (x_N , y_N) \big\} \subset \mathcal{X} \times \mathcal{Y} \enspace
\end{equation}

where $\mathcal{X}$ is the data space and $\mathcal{Y}$ is the set of outcomes that can be associated with an instance.

A neural network is a function $f_w: \mathcal{X} \mapsto \mathcal{Y}$ with trainable weights $w \in W$. Given training data, our model learns a set of weights $w^*$ that minimize a loss function $\text{L}$. Stochastic factors that impact the distribution of the learned weights $w^*$ at the end of training include both algorithm design choices (\texttt{ALGO}) that introduce noise to the training process and implementation choices (\texttt{IMPL}).

\begin{figure*}
\centering
\begin{small}
	\begin{sc}
    	\begin{subfigure}{0.29\linewidth}
    		\centering
        	\includegraphics[width=0.85\columnwidth]{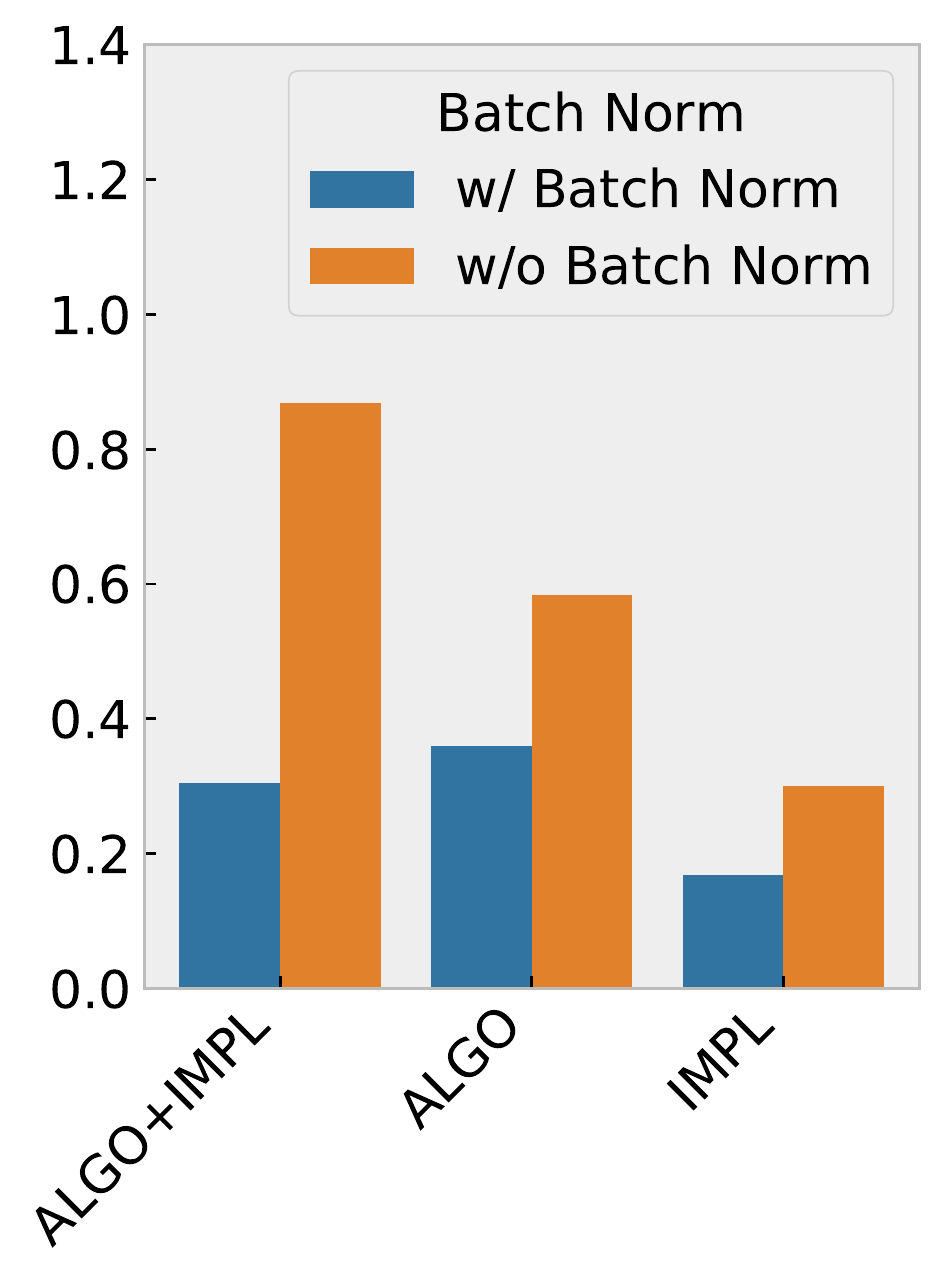}
            \caption{\textbf{STDDEV(Accuracy)}}
            \label{fig:smallcnn_bn_acc}
    	\end{subfigure}
    	\begin{subfigure}{0.29\linewidth}
    		\centering
            \includegraphics[width=0.85\columnwidth]{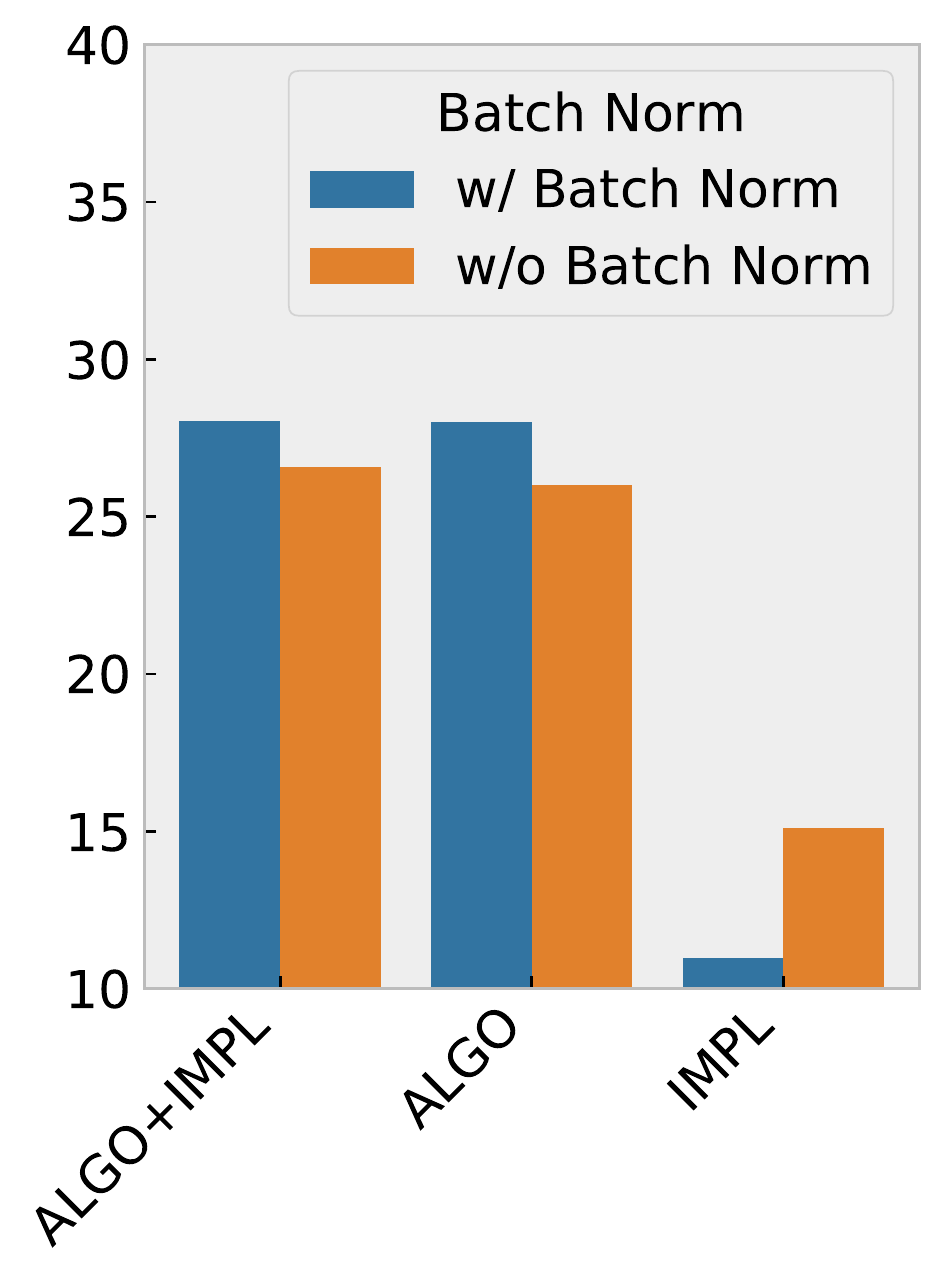}
            \caption{\textbf{Churn}}
            \label{fig:smallcnn_bn_churn}
    	\end{subfigure}
    	\begin{subfigure}{0.29\linewidth}
    		\centering
            \includegraphics[width=0.85\columnwidth]{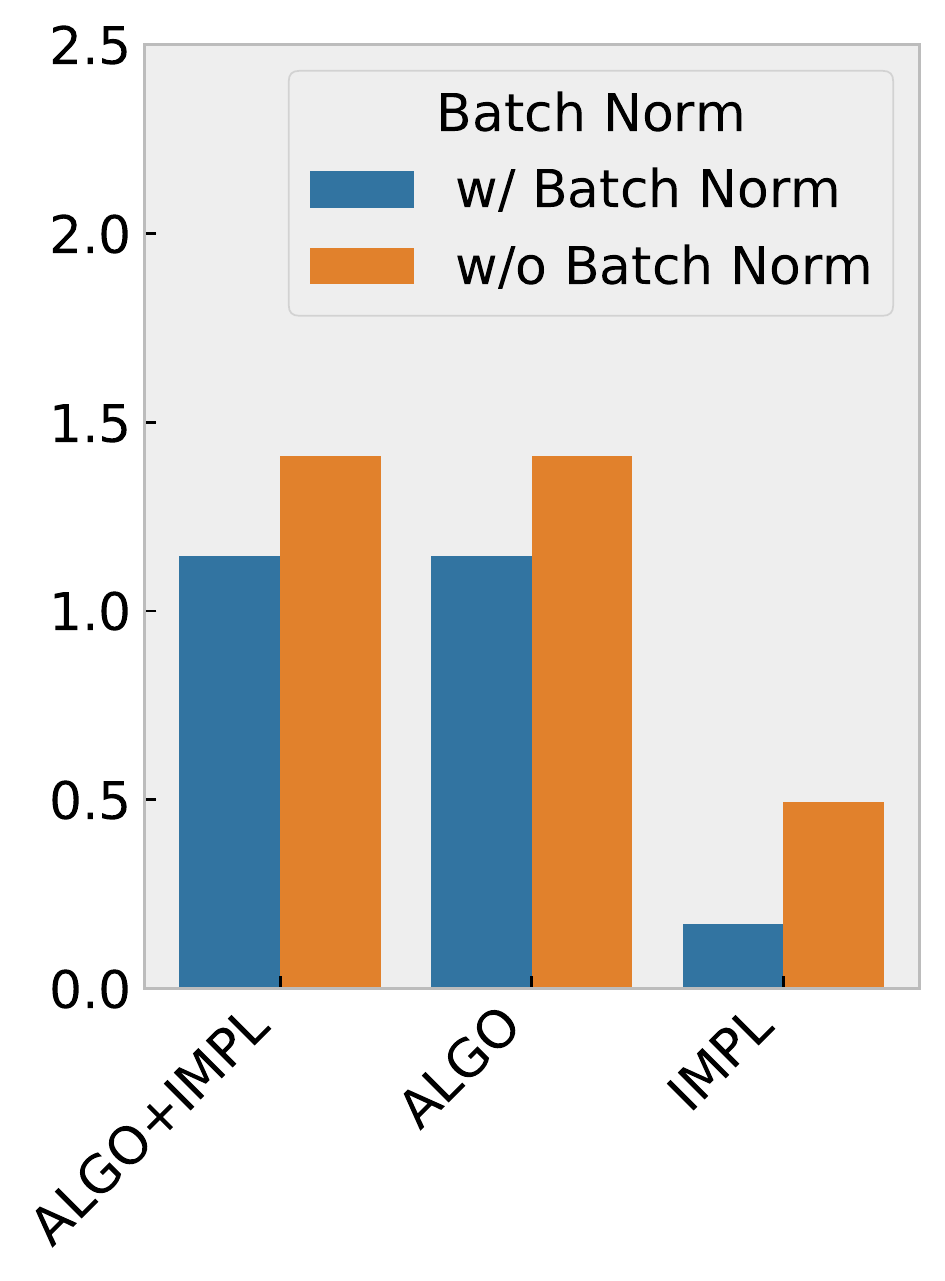}
            \caption{\textbf{L2 Norm}}
            \label{fig:smallcnn_bn_l2}
    	\end{subfigure}
	\end{sc}
\end{small}
\caption{Model design choices can amplify or curb impact of noise. Comparison of standard deviation of accuracy, prediction churn and l2 norm of 3-layer small CNN on CIFAR-10 dataset when trained with and without batch normalization in each layer.}
\label{fig:smallcnn_bn}
\end{figure*}

\textbf{Algorithmic Factors}~(\texttt{ALGO})~-~model design choices which are stochastic by design. For example, random initialization \citep{pmlr-v9-glorot10a,resnet}, data augmentation \citep{kukacka2017regularization,Hern_ndez_Garc_a_2018}, data shuffling ordering \citep{smith2018dont}, and stochastic layers \citep{Srivastava2014,hinton2012improving,pmlr-v28-wan13,10.5555/3104322.3104425,merity2017regularizing}. In Appendix~\ref{subsec:appendix_sources_randomness}, we include  a more detailed treatment of the widely-used model design choices that introduce stochasticity in DNN training.


\textbf{Implementation Factors}~(\texttt{IMPL})~-~noise introduced by software choices (e.g. Tensorflow \citep{tensorflow}, PyTorch \citep{pytorch}, cuDNN \citep{cudnn}) as well as hardware accelerators' architectures (e.g., modern GPU hardware designs \citep{pascal, volta, turing}). The following describes two typical scenarios causing implementation noises. 

\begin{itemize}
\itemsep0em
\item \textbf{Parallel Execution}~-~Popular general-purpose DNN accelerators (e.g., GPUs) leverage highly parallel execution for speed-up in execution. However, these sophisticated software-hardware designs for fine-grained massive parallelism typically aims to maximize resource utilization for execution speed and throughput \emph{rather} than output accuracy/precision. GPUs introduce stochasticity due to random floating-point accumulation ordering, which often cause inconsistent outputs between multiple runs due to the truncation of fraction part in floating point number in the accumulation procedure \citep{dab}.
\item \textbf{Input Data Shuffling and Ordering}~-~While input data shuffling induces algorithmic noise, it also induces implementation noise due to the different input ordering. Differences in input data ordering can result in different floating point accumulation orders for the reduction operations across data points which are often a overlooked source of implementation noise.
\end{itemize}


\subsection{Measures of Model Stability}\label{sec:measures_model_stability}

In this work, we are focused on measuring the impact of randomness on model stability, defined as \textit{ensuring that given the same experimental framework and tooling, the variation of the training outcome for given input dataset}. To this end, we evaluate the impact on both top-line metrics, but also more granular measures of model stability such as predictive churn, l2 norm and sub-group performance, as different measures of model stability. We briefly introduce each below.

\textbf{Churn}~(\texttt{churn})-~Predictive churn is a measure of predictive divergence between two models. In sensitive domains such as medicine, consistent individualized predictions are of paramount importance, as there can be severe costs for inconsistent model behavior with a risk to human life \citep{NAP13284}. Thus, understanding the factors that amplify churn is of considerable research interest with several different proposed definitions of predictive churn \citep{chen2020point,Shamir2020,Snapp2021}. We define churn between two models $f_1$ and $f_2$ as done by \citep{NIPS2016_dc5c768b} as the fraction of test examples where the predictions of two models disagree.:
\begin{align}
C(f_1, f_2) &= \mathbb{E}_{\mathcal{X}}\big[\mathbbm{1}_{\{\hat{\mathcal{Y}}_{x; f_1} \neq \hat{\mathcal{Y}}_{x; f_2}\}}\big]
\end{align}

where $\mathbbm{1}$ is an indicator function for whether the predictions by each model match. 

\textbf{L2 norm}~(\texttt{l2})~-~L2 norm of the trained weights $\| w^*_1 - w^*_2\|$  between $f_1$ and $f_2$ at the end of training indicates the divergence of each run in function space. We normalize the weight vector to a unit vector before computing l2 norm, for a consistent visualization scale across a variety of experiments.

\textbf{Standard Deviation of top-line and sub-group metrics}~(\texttt{stdev})~-~In addition to the standard deviation of top-1 test-set accuracy over independent runs, we measure deviation in sub-group performance as measured by sub-group error rate, false positive rate (FPR) and false negative rate (FNR). We compute standard deviation over 10 independent runs unless indicated otherwise.


\begin{table*}
    \caption{Test-set accuracy with standard deviation under each type of noise. We report the average of 10 models trained independently from scratch.}
    \label{table:accuracy_breakdown}
    \medskip
    \centering
  \scalebox{1.0}{
    \begin{tabular}{ccccc}
         \toprule
  \textbf{Hardware}  &  \textbf{Task}  &   \multicolumn{3}{c}{\textbf{Test Accuracy}}          \\
      &    & \texttt{ALGO+IMPL}     & \texttt{ALGO}       & \texttt{IMPL}    \\
    \midrule
 &   SmallCNN CIFAR-10   &\pctrange{62.28}{0.83}&\pctrange{61.44}{0.41}&\pctrange{61.61}{0.31} \\
 P100 &   ResNet18 CIFAR-10   &\pctrange{93.33}{0.14}&\pctrange{93.32}{0.13}&\pctrange{93.12}{0.11}\\
&    ResNet18 CIFAR-100  &\pctrange{73.37}{0.23}&\pctrange{73.42}{0.26}&\pctrange{73.36}{0.17} \\
       \midrule
    \midrule
&      SmallCNN CIFAR-10   &\pctrange{62.24}{0.64}&\pctrange{62.13}{0.85}&\pctrange{62.36}{0.16} \\
RTX5000 &  ResNet18 CIFAR-10   &\pctrange{93.34}{0.11}&\pctrange{93.44}{0.19}&\pctrange{93.13}{0.09} \\
&  ResNet18 CIFAR-100  &\pctrange{73.30}{0.16}&\pctrange{73.52}{0.15}&\pctrange{73.34}{0.24} \\
      \midrule
    \midrule
&      SmallCNN CIFAR-10   &\pctrange{62.03}{0.91}&\pctrange{62.35}{0.61}&\pctrange{61.69}{0.31} \\
V100  &  ResNet18 CIFAR-10   &\pctrange{93.32}{0.17}&\pctrange{93.44}{0.05}&\pctrange{93.41}{0.13}\\
&    ResNet18 CIFAR-100  &\pctrange{73.42}{0.25}&\pctrange{73.35}{0.14}&\pctrange{73.41}{0.28}\\
 &   ResNet50 ImageNet   &\pctrange{76.58}{0.10}&\pctrange{76.61}{0.10}&\pctrange{76.60}{0.05} \\
    \bottomrule 
    \end{tabular}
  }
\end{table*}

\subsection{Experimental Setup}

We conduct extensive experiments across large-scale datasets (CIFAR-10 and CIFAR-100 \citep{Krizhevsky09learningmultiple}, ImageNet \citep{IRSVRC} and CelebA \citep{celeba}) and widely-used networks including ResNet-18 and ResNet-50 \citep{resnet}, DenseNet-121 and DenseNet-201 \citep{densenet}, Inception-v3 \citep{szegedy2015rethinking}, MobileNet \citep{mobilenetv2}, EfficientNet \citep{tan2020efficientnet}, three-layer small CNN and six-layer medium CNN (Appendix~\ref{sec:appendix:smallcnn}). For all the experiment variants with the exception of ImageNet, we report the average performance metric over 10 models independently trained from scratch. For ImageNet, given the higher training cost, we report average performance across 5 independent trains.  Table~\ref{table:accuracy_breakdown} includes the baseline accuracy given each dataset/model combination we train. A detailed description of training methodology for each dataset and model architecture combination is included in Appendix~\ref{sec:appendix:training_methodology}. We preserve the same hyperparameter choices across hardware types and use Tensorflow \citep{tensorflow} 2.4.1, CUDA 11, and cuDNN \citep{cudnn} 8 for all the experiments.

\textbf{GPU}~-~we evaluate NVIDIA P100 with an older Pascal architecture \citep{pascal} and later generations  NVIDIA V100 \citep{volta}, Nvidia RTX5000 and T4 \citep{turing} with Volta and Turing architecture respectively. Our choice of GPUs allows us to evaluate the impact of different levels of parallelism, as P100, V100, RTX5000, and T4 GPU are each equipped with 3584, 5120, 3072, and 2560 CUDA Cores for floating point computation, respectively. In addition, we compare GPUs with and without Tensor Cores accelerators by evaluating both Pascal and Turing architectures. GPU generations with Turing architectures have multiple dedicated matrix multiplication units called Tensor Cores to provide massive computation throughput. 

\textbf{TPU}~-~A TPU \citep{jouppi2017indatacenter} is a custom ASIC which differs from GPUs that employ arithmetic logic unit (ALU) as the basic building blocks to offer massive parallel computation. In contrast, TPUs leverage systolic arrays \citep{systolicarray} in matrix unit (MXU) to provide massive computation throughput with a single-threaded, deterministic computation model. Thus, TPUs are designed to be deterministic, which differs from the time-varying optimizations of CPUs and GPUs such as caches, out-of-order-execution, multithreading, MIMD/SIMD and prefetching, etc.

We benchmark four key experimental variants which allows us to independently measure the impact of both algorithm (\texttt{ALGO}) and implementation (\texttt{IMPL}) factors on downstream model performance:

\textbf{Both Algorithm + Implementation noise}~-~(\texttt{ALGO + IMPL}). Here, we do not control for either algorithmic or implementation factors that introduce randomness. This is the default setting of the model training procedure.

\textbf{Only Algorithm noise}~-~(\texttt{ALGO}). We measure the impact of stochastic algorithmic factors by fully controlling all noise introduced by tooling. The implementation noise controlling feature is supported by many prevalent deep learning frameworks such as Tensorflow \citep{tensorflow} and Pytorch \citep{pytorch}. Note that controlling implementation noise is far from free (Section~\ref{sec:result:overhead}).

\textbf{Only Implementation noise}~-~(\texttt{IMPL}). We measure the impact of implementation noise by using a fixed random seed for all stochastic algorithm factors. This results in deterministic weights initialization, data augmentation and batch shuffling.

\textbf{Control}~-~This \texttt{Control} variant both sets a fixed random seed to control algorithmic noise and uses software patches to eliminate implementation noise. 

\section{Results: Characterizing the Impact of Randomness}\label{sec:results_impact_randmness}
In this section we address the following questions: \textbf{1)} How do implementation and algorithmic noise contribute to system level noise? \textbf{2)} How do both impact model stability? \textbf{3)} How does varying choices of hardware, low-level vendor libraries and architecture impact the level of noise in the system, and \textbf{(4)} Why are certain model design choices far more sensitive to noise? 

\begin{figure*}[t]
\centering
    \includegraphics[width=0.6\linewidth]{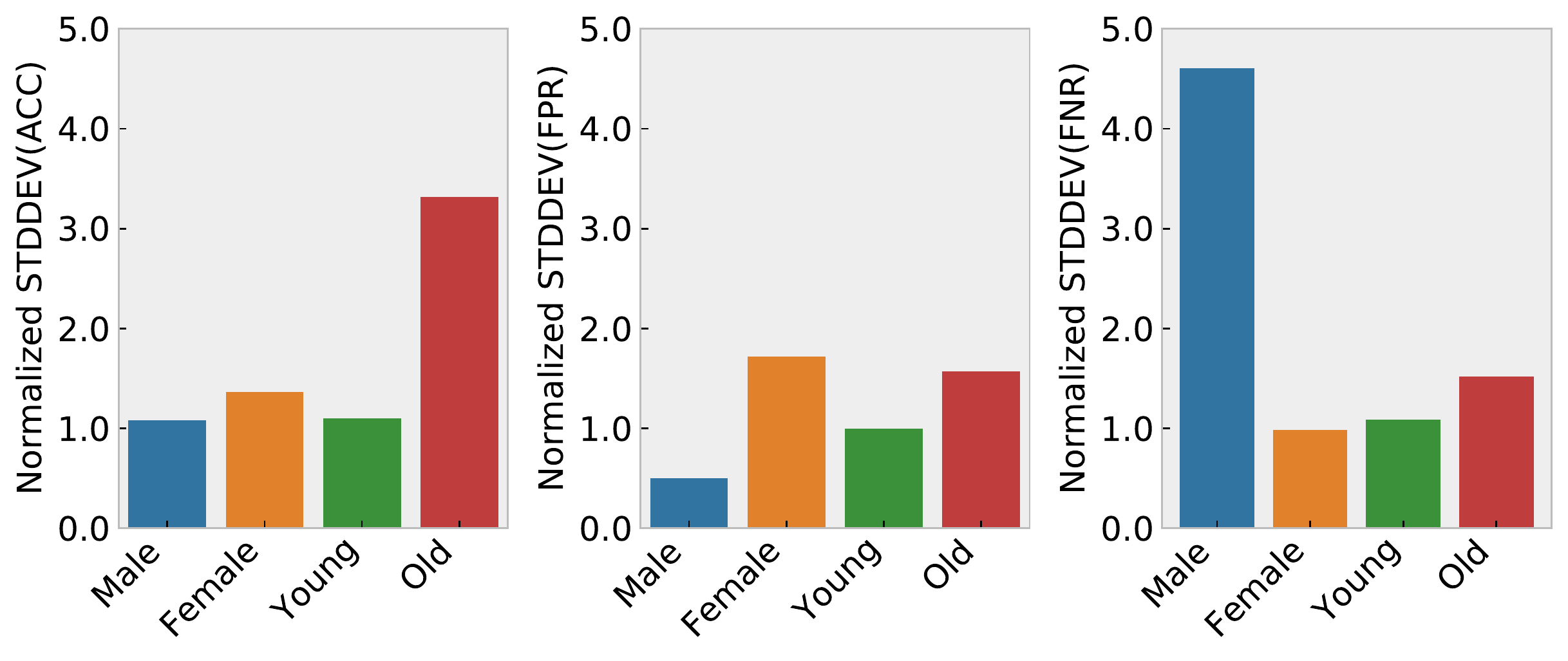}
    \caption{STDDEV(Accuracy) of each sub-group of ResNet18 trained on CelebA dataset using V100. Y axes is normalized against corresponding metric of overall dataset. Noise is disproportionately impacting \textit{Old} and \textit{Male} sub-group as these sub-groups have fewer data points for the positive class.}
    \label{fig:noise_impact_group_with_less_datapoints}
\end{figure*}
\begin{table}
    \caption{Data points distribution in CelebA dataset}
    \label{tab:celeba_data_distribution}
    \centering
    \scalebox{0.99}{
    \begin{tabular}{lcccc}
    \toprule
    & Male & Female & Young & Old \\
    \midrule
    Positive Data Points & \textbf{1387 (0.8\%)} & 22880 (14.1\%) & 20230 (12.4\%) & \textbf{4037 (2.5\%)} \\
    Negative Data Points & 66874 (41.1\%) & 71629 (44.0\%) & 106558 (65.5\%) & 31945 (19.6\%) \\
    \bottomrule
    \end{tabular}
    }
\end{table}

\subsection{Impact of Randomness on Top-Line Metrics}\label{sec:result:topline-metric}

\begin{figure*}
\centering
\begin{small}
	\begin{sc}
    	\begin{subfigure}{0.3\linewidth}
    		\centering
        	\includegraphics[width=0.99\columnwidth]{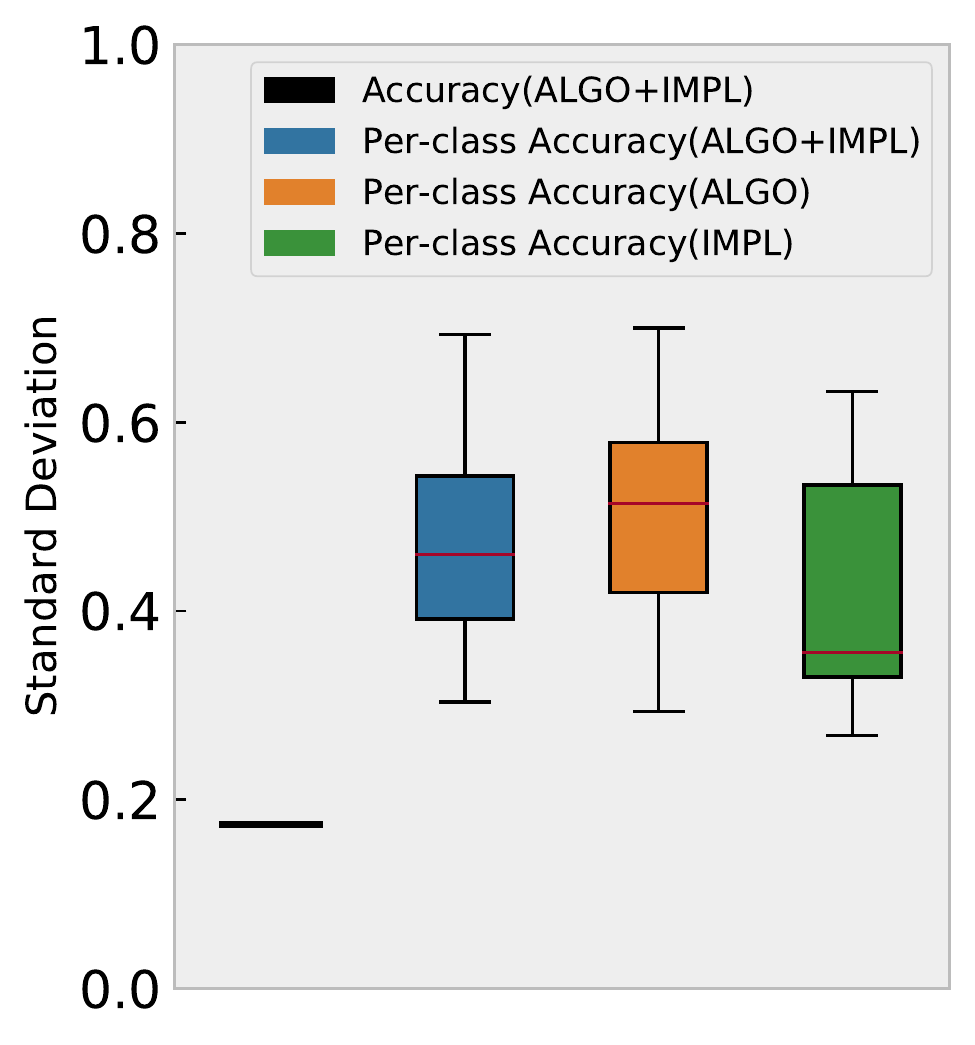}
            \caption{CIFAR-10}
            \label{fig:perclass_variance_cifar100}
    	\end{subfigure}
    	\begin{subfigure}{0.3\linewidth}
    		\centering
            \includegraphics[width=0.99\columnwidth]{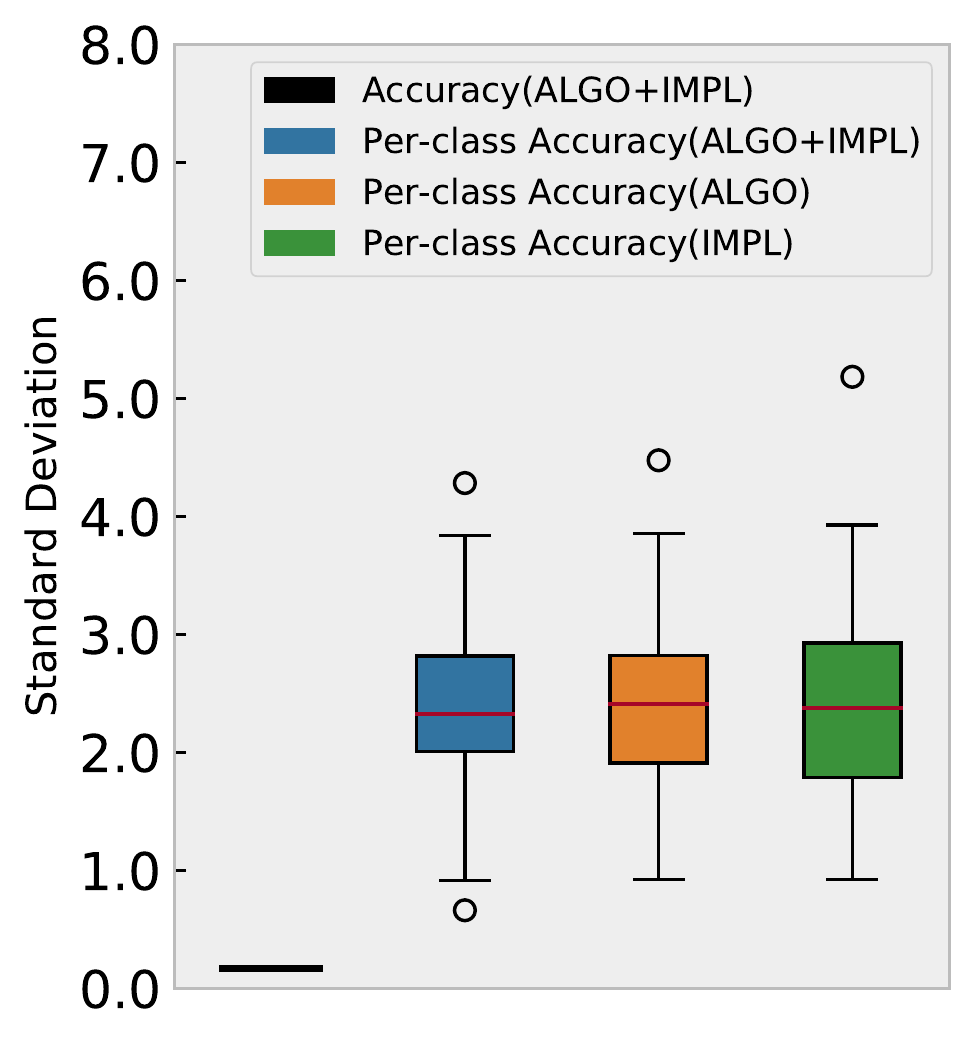}
            \caption{CIFAR-100}
            \label{fig:perclass_variance_cifar10}
    	\end{subfigure}
	\end{sc}
\end{small}
\caption{Per-class accuracy variance vs. overall accuracy variance of ResNet18 trained on V100 under different factors of noise. Per-class accuracy variance is up to \textbf{Left:} 4X larger for Cifar-10 and \textbf{Right:} 23X larger CIFAR-100 than overall accuracy.}
\label{fig:perclass_variance_cifar}
\end{figure*}

\textbf{Top-1 Accuracy} Across all experiments, we observe small variance in Top-1 accuracy. In Table~\ref{table:accuracy_breakdown}, the maximum standard deviation in accuracy is 0.91\% for the small cnn trained on CIFAR-10, and the minimum standard deviation is 0.05\% for ResNet-10 trained on ImageNet. Top-line metrics do not differ substantially between algorithmic and implementation factors, we observe there is less than 1\% standard deviation between \texttt{ALGO}, \texttt{IMPL} and \texttt{ALGO+IMPL} aross all datasets and networks.
We observe a maximum absolute difference of 0.84\% on the small cnn network trained on CIFAR-10. 

\textbf{Model Stability Metrics} A closer inspect of \texttt{l2}, \texttt{churn} and \texttt{stdev} measures in Figure~\ref{fig:types_noise_v100} show that both \texttt{ALGO} and \texttt{IMPL} factors create significant levels of model instability across each of these measures. While for most networks and measures, \texttt{ALGO} contributes higher levels of instability relative to \texttt{IMPL} factors, this is not always a pronounced gap. For example, on ResNet-50 ImageNet, the impact of predictive churn of \texttt{IMPL} factors is $14.68\%$ versus \texttt{ALGO} factors is $14.89\%$. Our results show that \texttt{IMPL} can be a significant source of non-determinism accumulated along the trajectory of model training procedure. Due to the non-linearities in deep neural network training, simply removing a single source of noise cannot effectively reduce the level of uncertainty of trained models.

Combined sources of noise (\texttt{ALGO} + \texttt{IMPL}) are a non-additive combination of individual factors. For example, the impact of (\texttt{ALGO} + \texttt{IMPL}) factors on churn for ResNet-18 and ResNet-50 is on par or only slightly higher than the impact of \emph{only} \texttt{IMPL} or \texttt{ALGO} noise. The lack of an additive relationship between different sources of noise suggests there is an upper bound in what level of overall system noise is possible.


\textbf{The role of model design choices} In Figure~\ref{fig:types_noise_v100}, we observe pronounced amplification of noise in the small CNN relative to ResNet-18 for CIFAR-10 with far higher \texttt{stdev}, \texttt{churn} and \texttt{l2} for all sources of noise. The small CNN is the only architecture we benchmark without batch normalization (BN) \citep{Ioffe2015}, a standard technique for stabilizing training \citep{tessera2021gradients}. To understand the role of model design choices at curbing or amplifying overall noise in the system, we  evaluate the impact to training with and without BN. We compare the small CNN trained without BN to the same architecture trained with BN. In Figure~\ref{fig:smallcnn_bn}~(a), we show that BN has a pronounced impact with a decline in the \texttt{stddev} of the accuracy from $0.86\%$ without BN to a much small $0.30\%$ with BN.

We note that architecture appears to play a larger role than dataset in the amplification or curbing of system noise. For example, in Figure~\ref{fig:types_noise_v100} the difference in standard deviation between small CNN (0.91\%) and ResNet-18 (0.17\%) is far larger than the difference between ResNet-18 trained on CIFAR-10 (0.17\%) vs the same architecture trained on CIFAR-100 (0.25\%).

\begin{figure*}
\centering
\begin{small}
	\begin{sc}
    	\begin{subfigure}{0.23\linewidth}
    		\centering
        	\includegraphics[width=0.99\columnwidth]{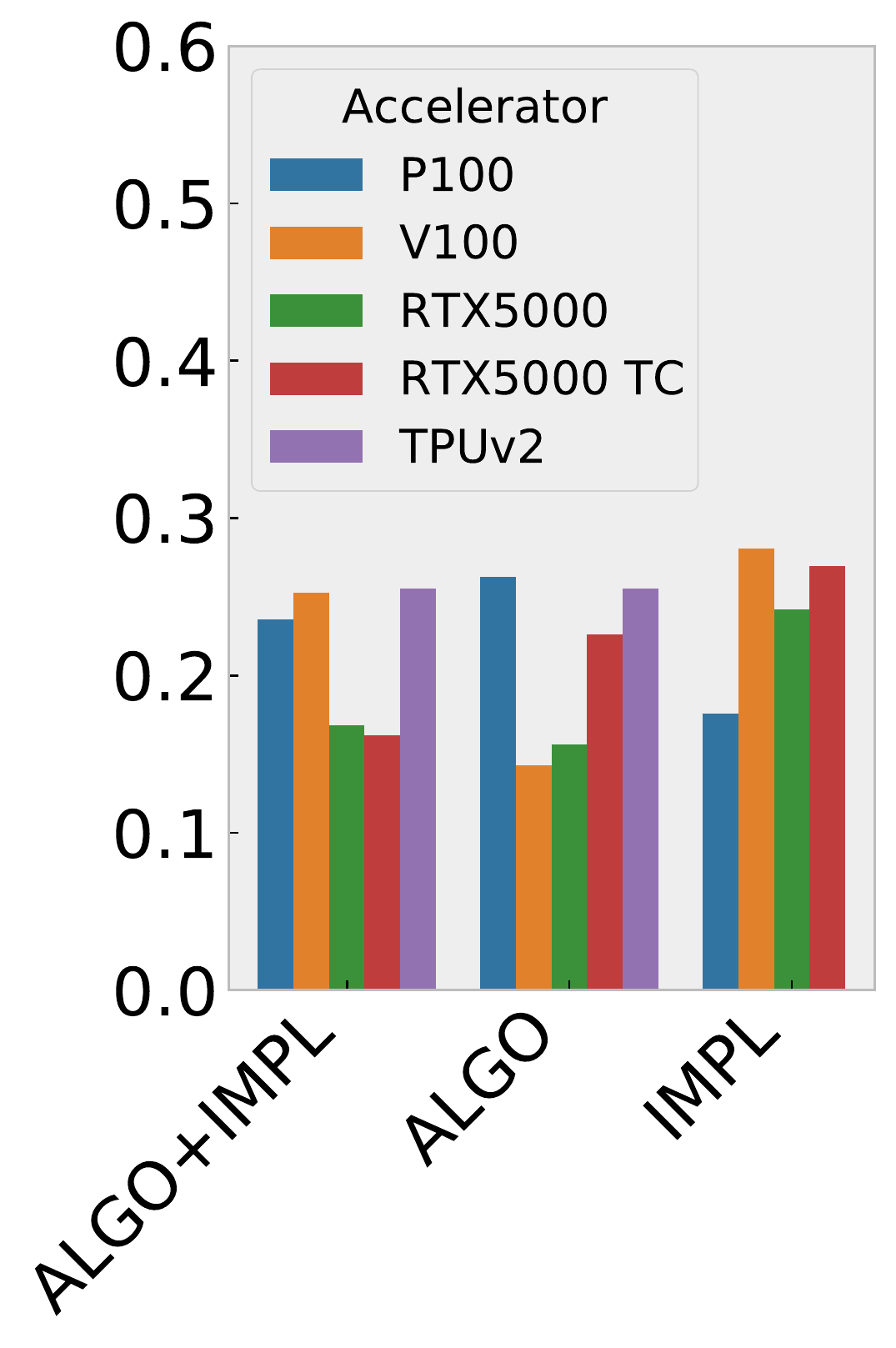}
            \caption{\textbf{STDDEV(Accuracy)}}
            \label{fig:cifar100_variance_over_gpus}
    	\end{subfigure}
    	\begin{subfigure}{0.23\linewidth}
    		\centering
            \includegraphics[width=0.99\columnwidth]{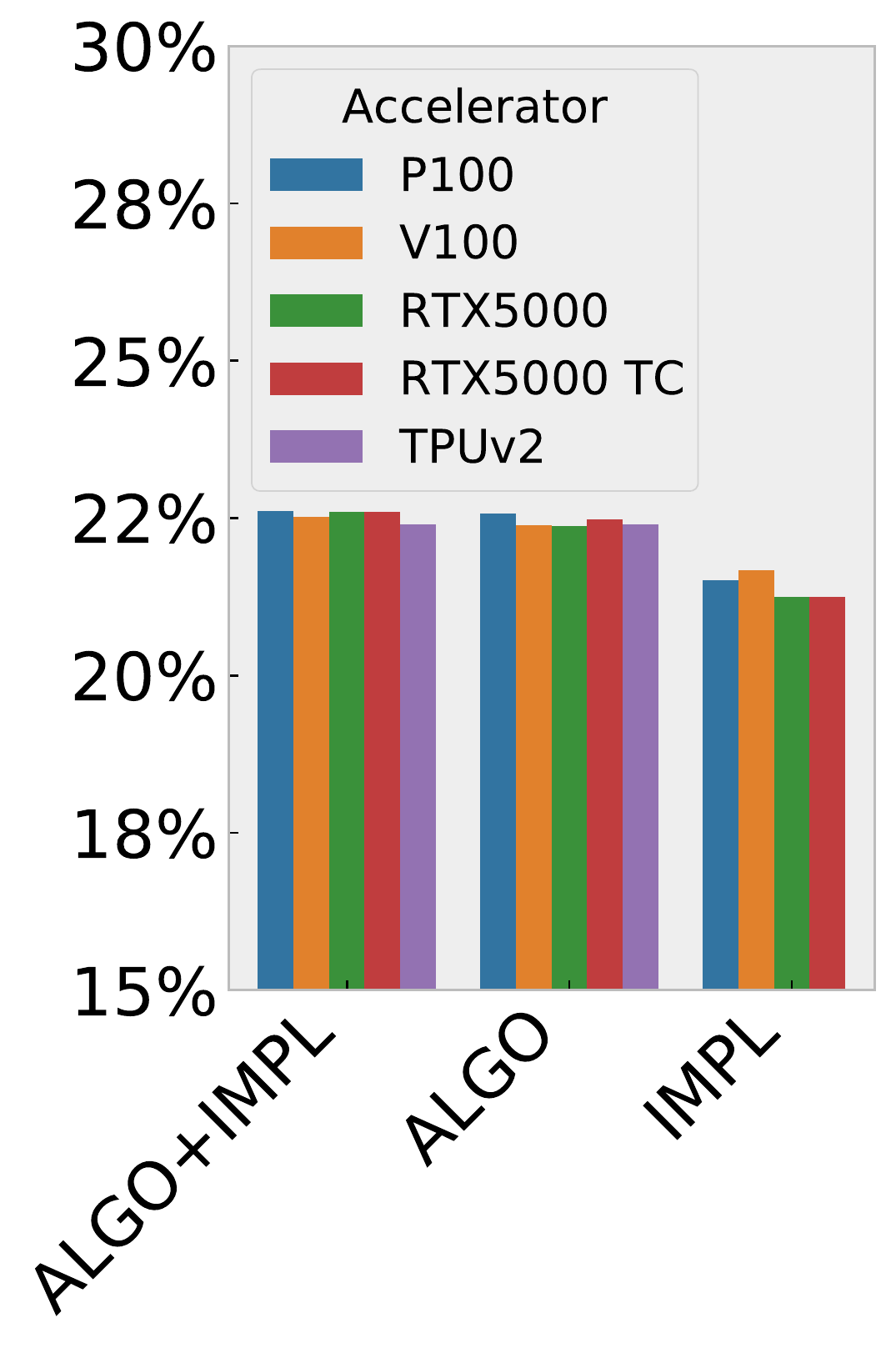}
            \caption{\textbf{Churn}}
            \label{fig:cifar100_churn_over_gpus}
    	\end{subfigure}
    	\begin{subfigure}{0.23\linewidth}
    		\centering
            \includegraphics[width=0.99\columnwidth]{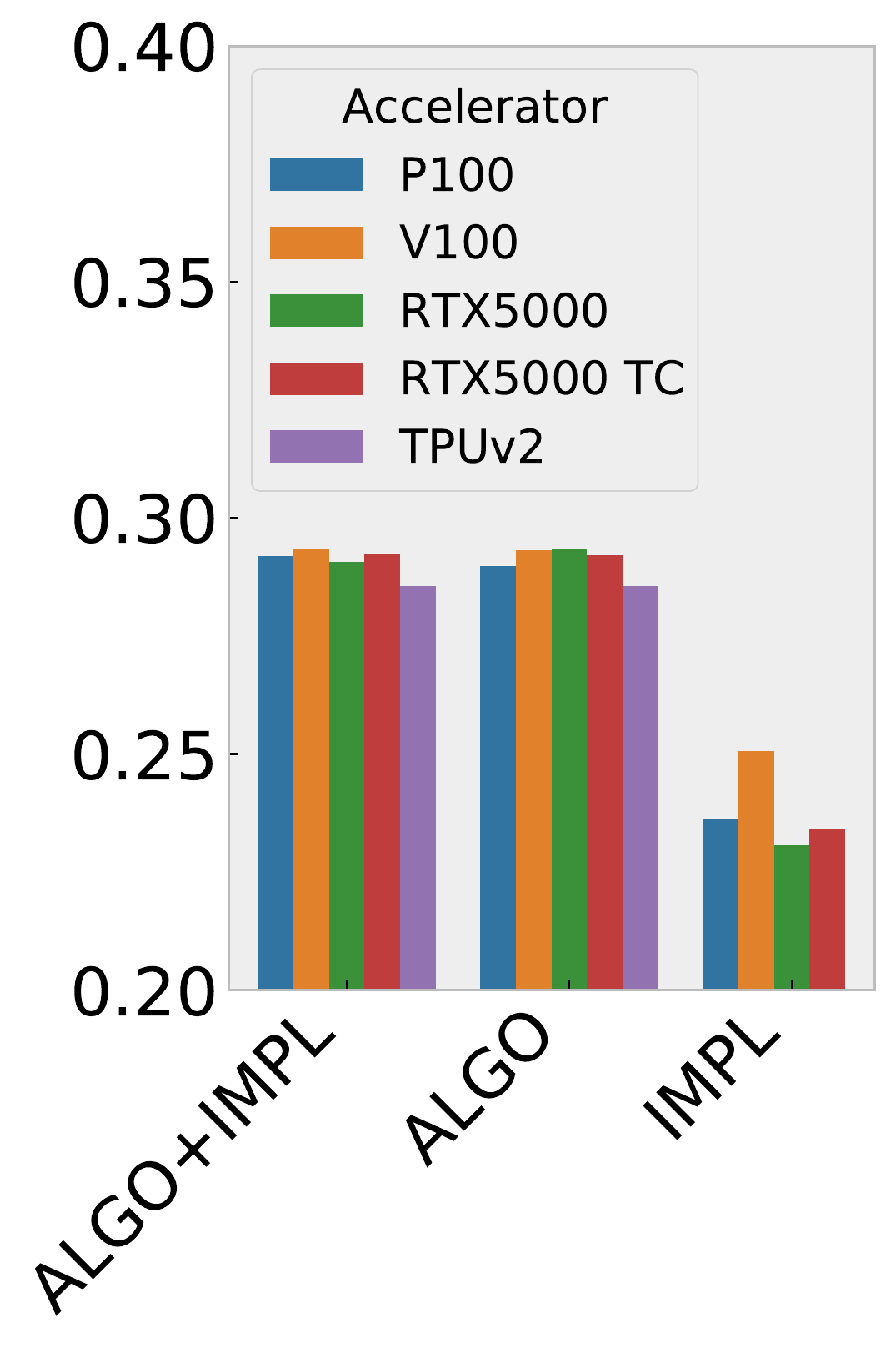}
            \caption{\textbf{L2 Norm}}
            \label{fig:cifar100_l2_over_gpus}
    	\end{subfigure}
	\end{sc}
\end{small}
\caption{Comparison of standard deviation of accuracy, prediction churn and l2 norm of ResNet18 on CIFAR-100 dataset between different training accelerators.}
\label{fig:cifar100_noise_over_gpus}
\end{figure*}

\begin{figure*}
  \centering
  \begin{minipage}[t]{0.3\linewidth}
    \centering \includegraphics[height=5cm]{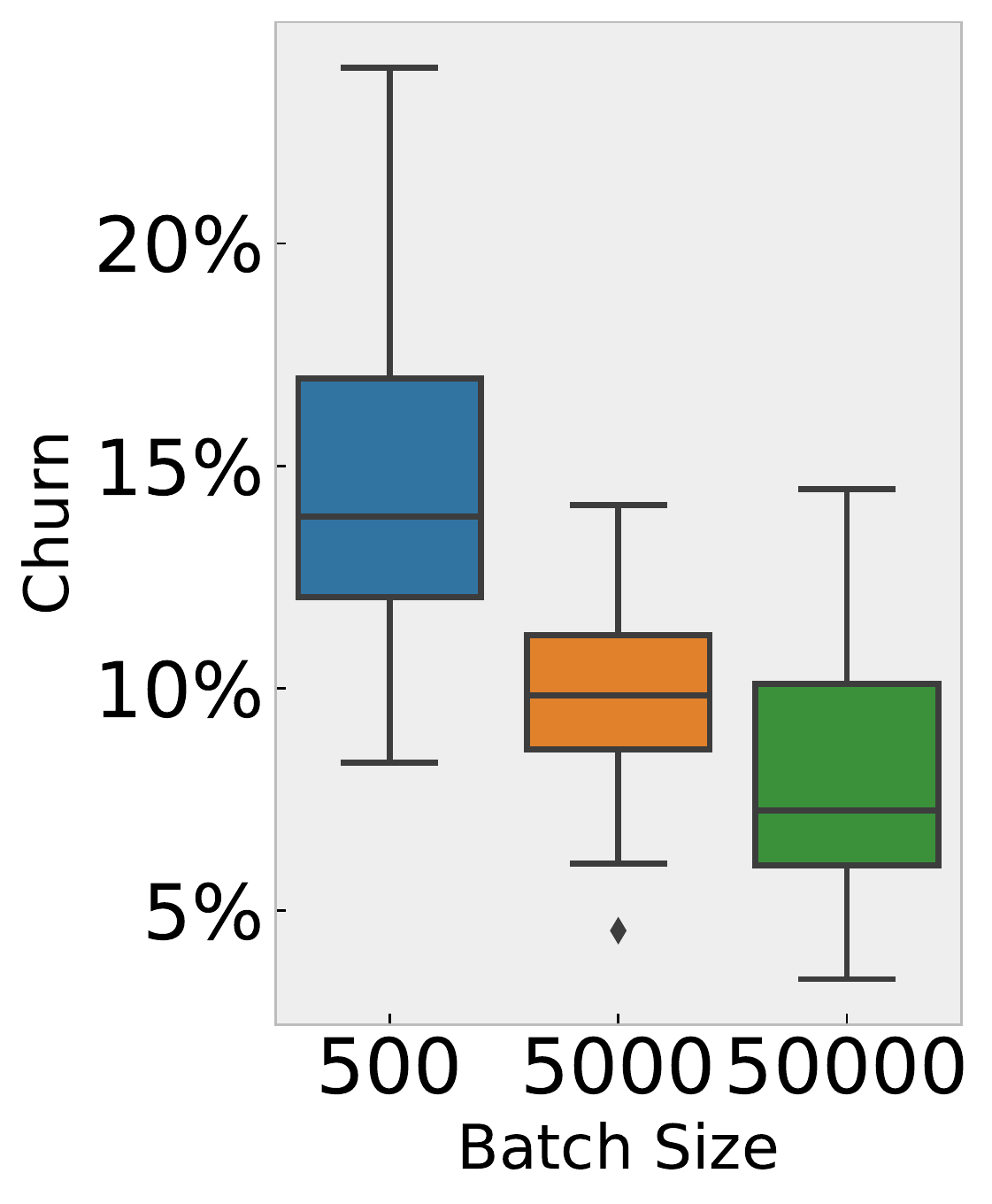}
    \caption{Data input order introduces additional non-determinism on TPU.}
    \label{fig:ordering-nondeterminism}
  \end{minipage}
  \hfill
  \begin{minipage}[t]{0.65\linewidth}
    \centering  \includegraphics[height=5cm]{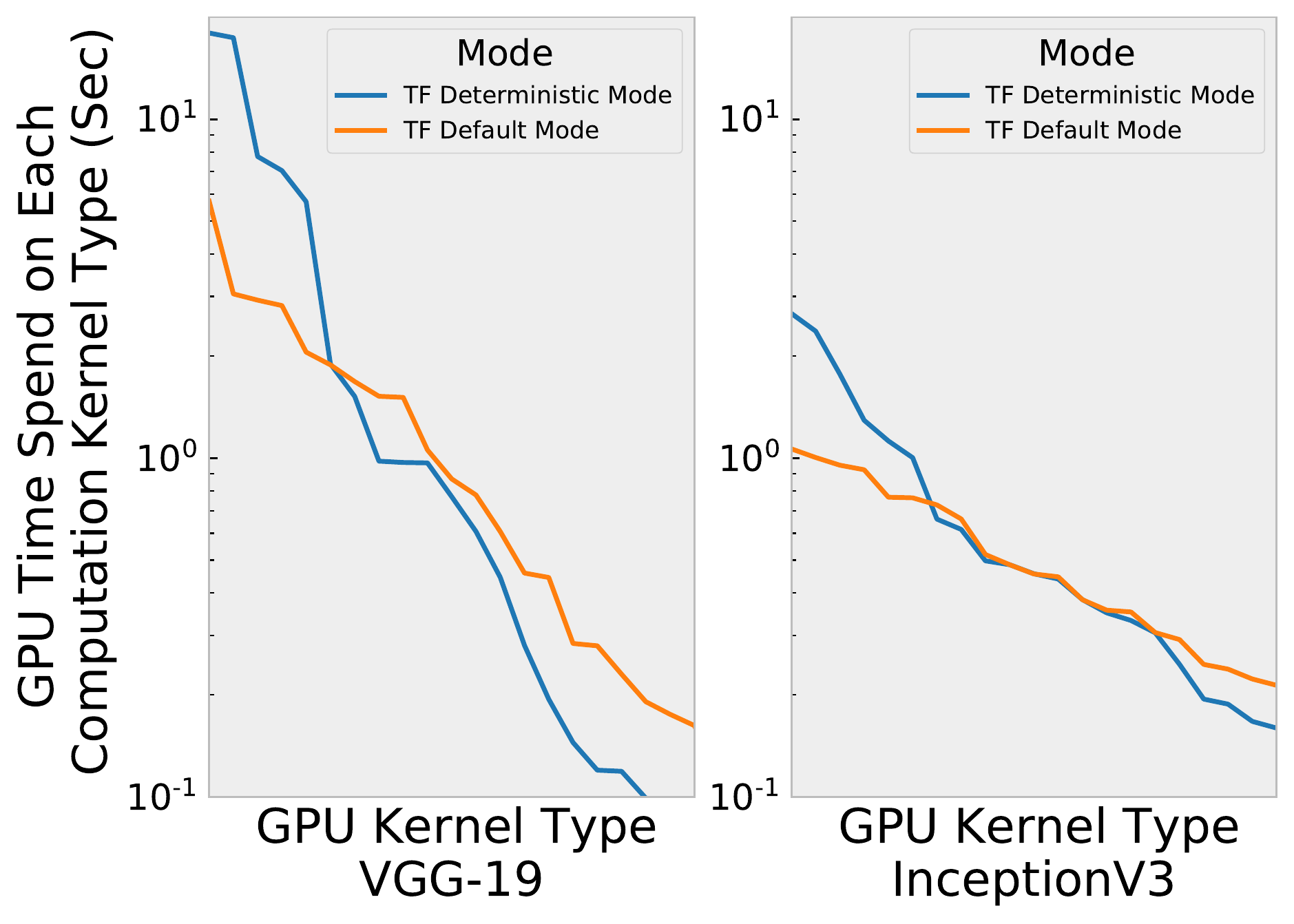}
    \caption{Top-20 GPU kernels cumulative runtime comparison (Top-1 is on the left side). X-axis indicates the different type of kernels scheduled on GPU. Y-axis is in log scale indicates the cumulative time spend on each type of GPU kernel.}
    \label{fig:overhead_overview}
  \end{minipage}
  \vskip -0.3in
\end{figure*}


\subsection{Impact of Randomness on Sub-Group Performance}\label{sec:result:sub-group-performance}

\textbf{How does noise impact sub-group performance?} We decompose top-line metrics along class label dimension on CIFAR-10/100 dataset \citep{Krizhevsky09learningmultiple} and CelebFaces Attributes (CelebA) dataset \cite{celeba}. In Figure~\ref{fig:perclass_variance_cifar}, we train models on CIFAR-10/100 under \texttt{ALGO+IMPL}, \texttt{ALGO}, and \texttt{IMPL} respectively. We observe high variance of per-class accuracy of \texttt{ALGO} and \texttt{IMPL} group similar to models trained under \texttt{ALGO+IMPL}. It is clear that removing partial source of noise does not effectively improve model stability. The maximum per-class standard deviation of accuracy is 4X and 23X on CIFAR-10 and CIFAR-100 dataset compared to standard deviation of top-1 accuracy. 

CelebA is a dataset of celebrity images where each image is associated with 40 binary labels identifying attributes such as hair color, gender, and age. Our goal is to understand the implications of noise on model bias and fairness considerations. Thus, we focus attention on two protected unitary attributes \texttt{Male}, \texttt{Female} and \texttt{Young} and \texttt{Old}. In Figure~\ref{fig:noise_impact_group_with_less_datapoints}, we can see that (\texttt{ALGO+IMPL}) noise is resulting unstable metrics on underrepresented \textit{Male} and \textit{Old} subgroups leading to disproportionate high-variance up to 3.3X on standard deviation on accuracy of \textit{Old} group and 4.6X standard deviation on FNR of \textit{Male} group. Thus, We conclude that even if the top-line metric variation is small enough, noise still imposes disproportionate high variance on dis-aggregated metrics. 

\textbf{Why are certain parts of the data distribution more sensitive to noise?} We observe a correlation between underrepresented sub-groups and the sub-groups with the most pronounced impact in variance. In Figure~\ref{fig:noise_impact_group_with_less_datapoints}, the classes disproportionately impacted \textit{Male} and \textit{Old} as they are heavily underrepresented in the training dataset with 0.8\% and 2.5\% positive labels as a fraction of the entire dataset (see Table.~\ref{tab:celeba_data_distribution}). This suggests stochasticity disproportionately impact features in the long-tail of the dataset.

\subsection{How does noise level vary across hardware types?}\label{sec:result:hardware-noise}
We trained ResNet-18 on CIFAR-10 and CIFAR-100 dataset using different accelerators including GPU using CUDA Cores (P100, V100, RTX5000), GPU using Tensor Cores (RTX5000 TC), and TPUv2-8 chip \citep{tpuv2}. For each hardware except TPU, we measure model divergence under each variant of source of noise (i.e. \texttt{ALGO+IMPL}, \texttt{ALGO}, and \texttt{IMPL}).

\begin{figure*}[ht!]
	\centering
	\vskip 0.15in
    \begin{small}
    \begin{sc}
    \begin{subfigure}{0.49\linewidth}
        \begin{subfigure}{0.32\linewidth}
		    \centering \includegraphics[width=0.99\linewidth]{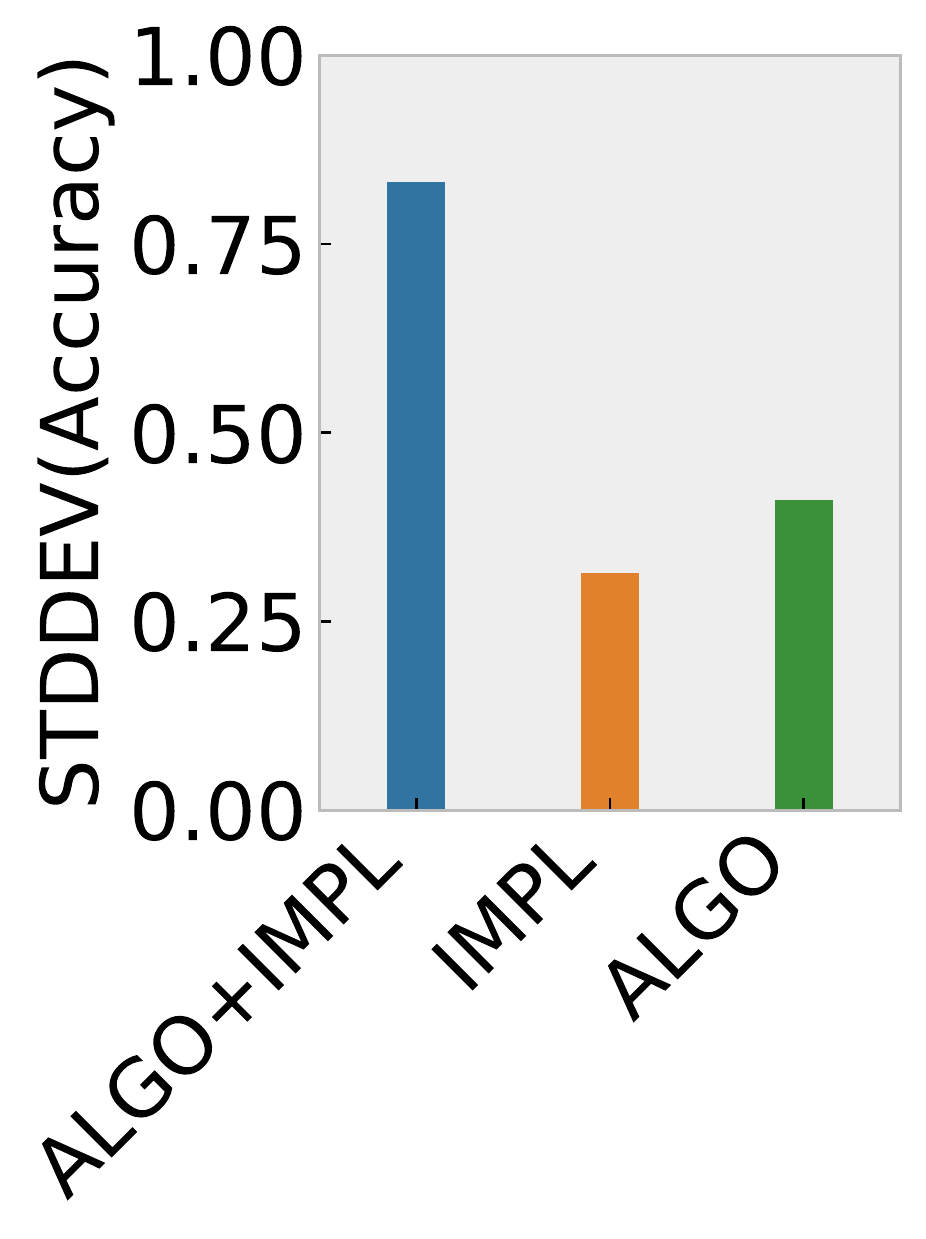}
        \end{subfigure}
        \hspace{-0.5em}
        \begin{subfigure}{0.32\linewidth}
        	\centering \includegraphics[width=0.99\linewidth]{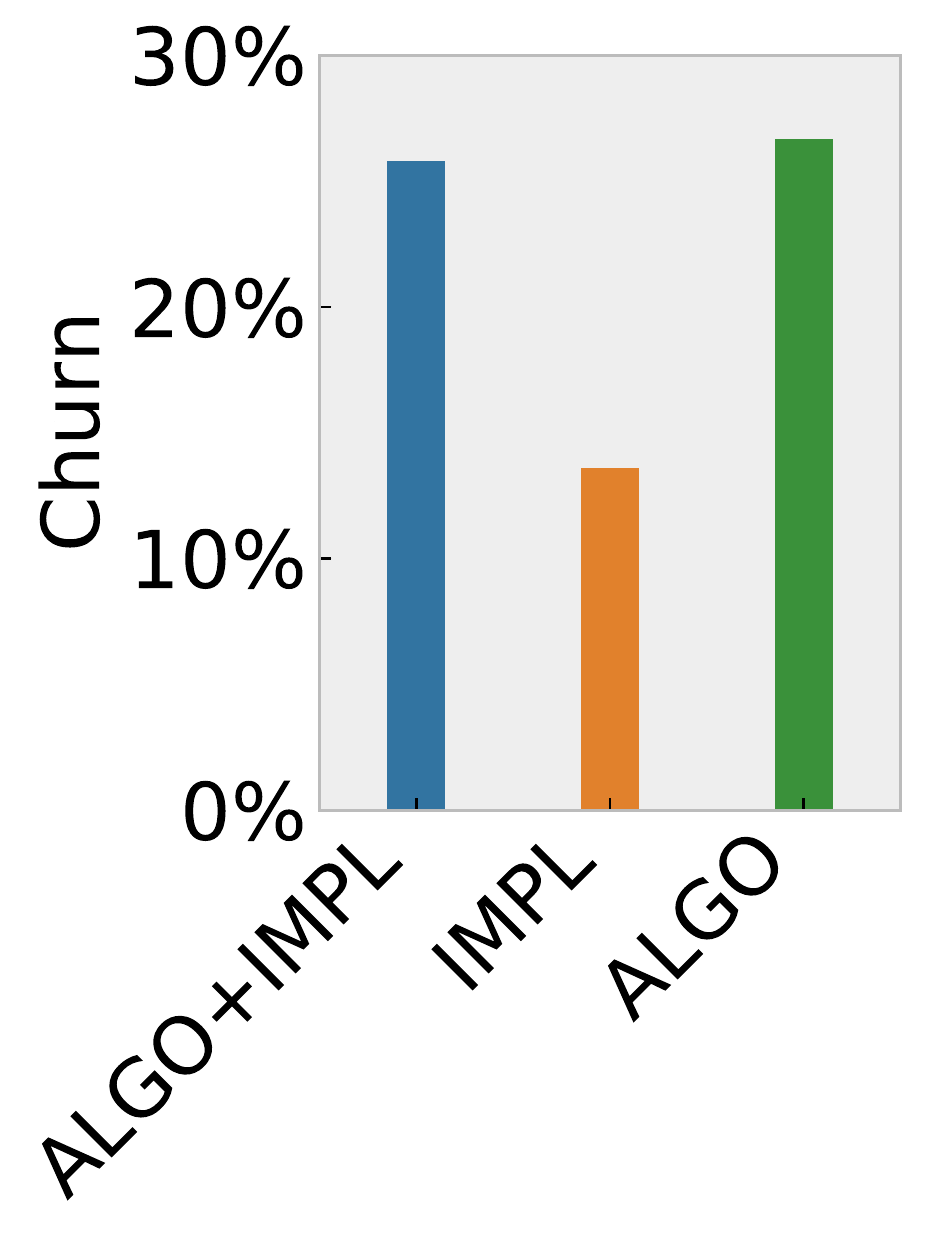}
        \end{subfigure}
        \hspace{-0.5em}
        \begin{subfigure}{0.32\linewidth}
        	\centering \includegraphics[width=0.99\linewidth]{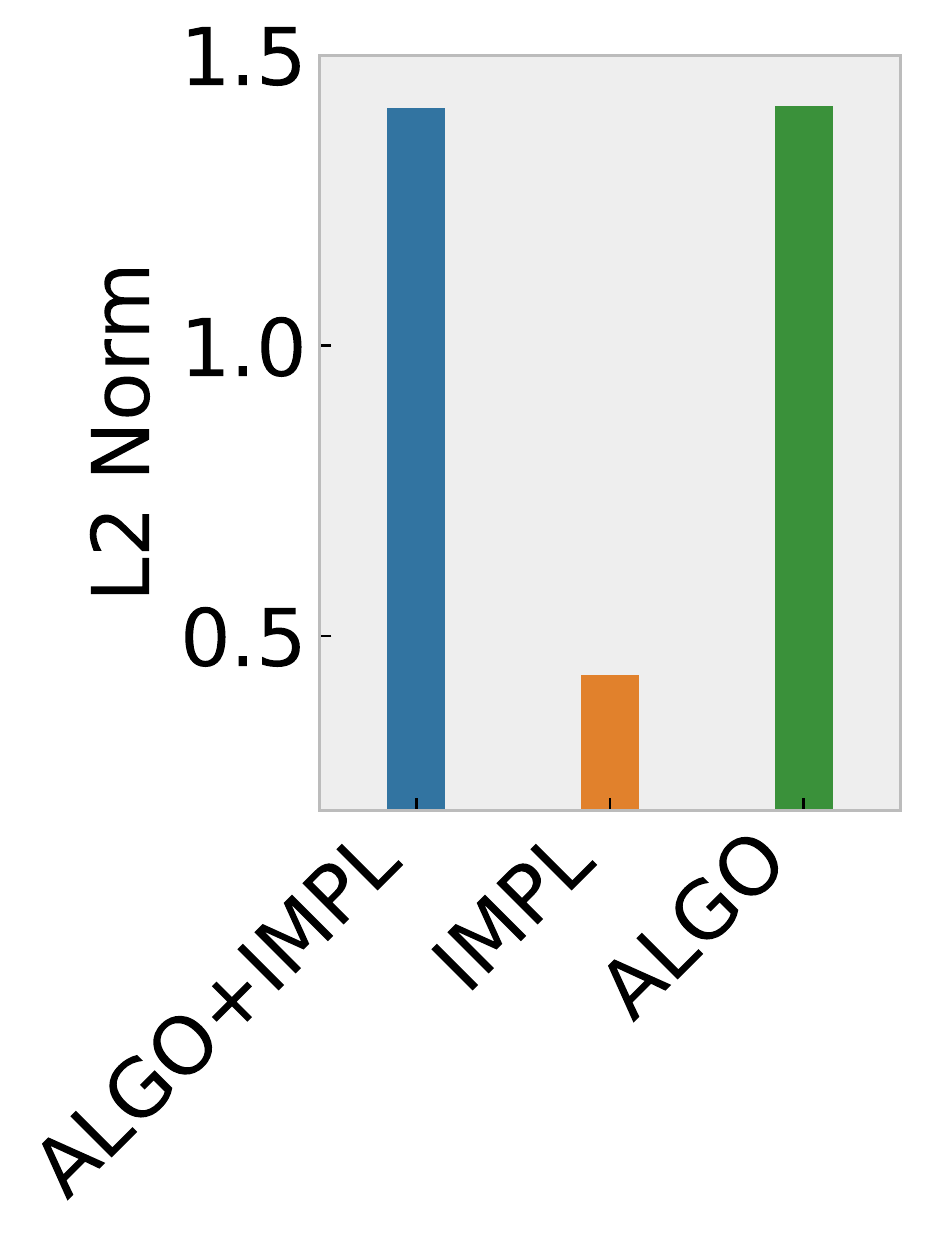}
        \end{subfigure}
        \caption{SmallCNN CIFAR-10}
    \end{subfigure}
    \\
    \begin{subfigure}{0.49\linewidth}
        \begin{subfigure}{0.32\linewidth}
    		\centering \includegraphics[width=0.99\linewidth]{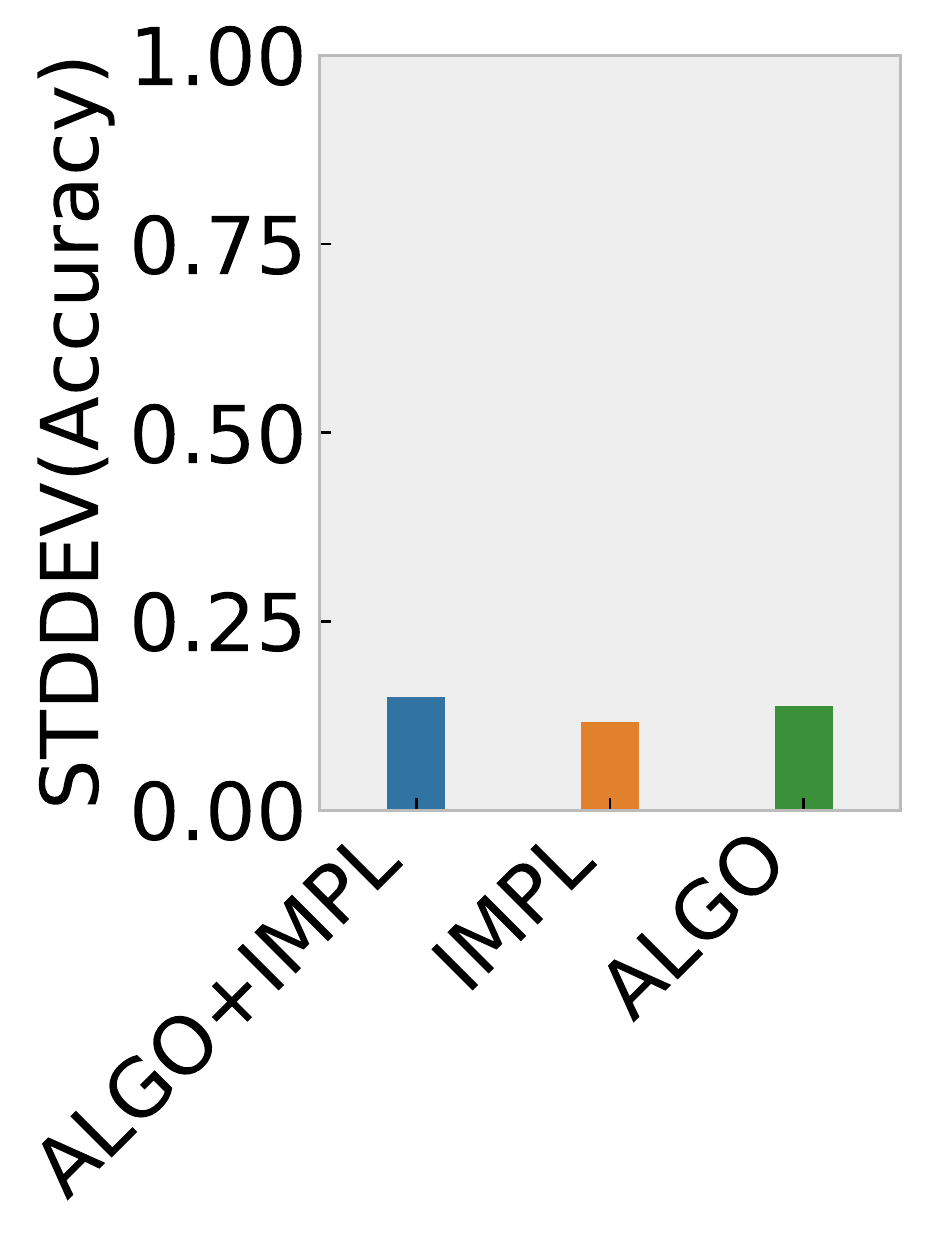}
        \end{subfigure}	
        \hspace{-0.5em}
        \begin{subfigure}{0.32\linewidth}
        	\centering	\includegraphics[width=0.99\linewidth]{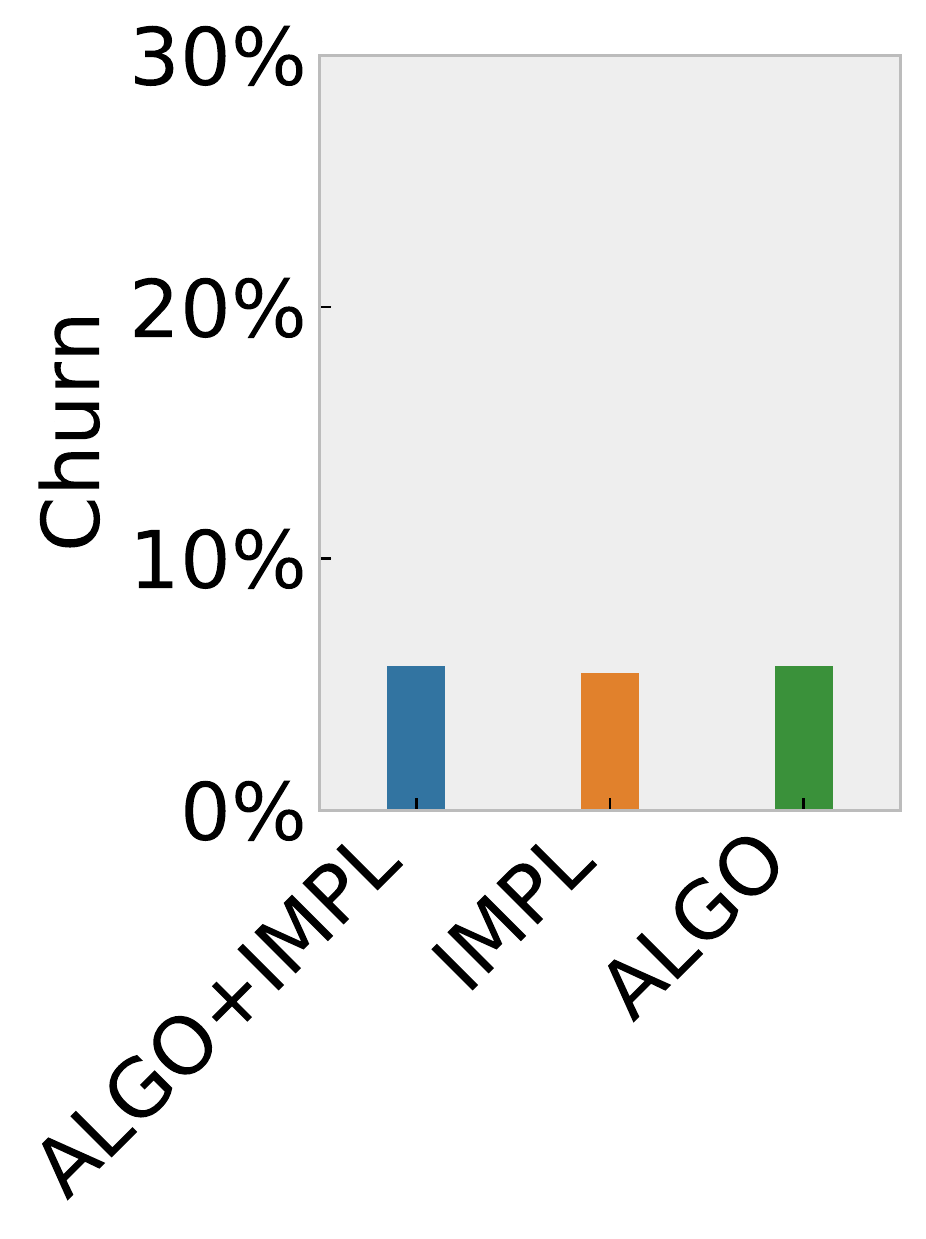}
        \end{subfigure}
        \hspace{-0.5em}
        \begin{subfigure}{0.32\linewidth}
        	\centering \includegraphics[width=0.99\linewidth]{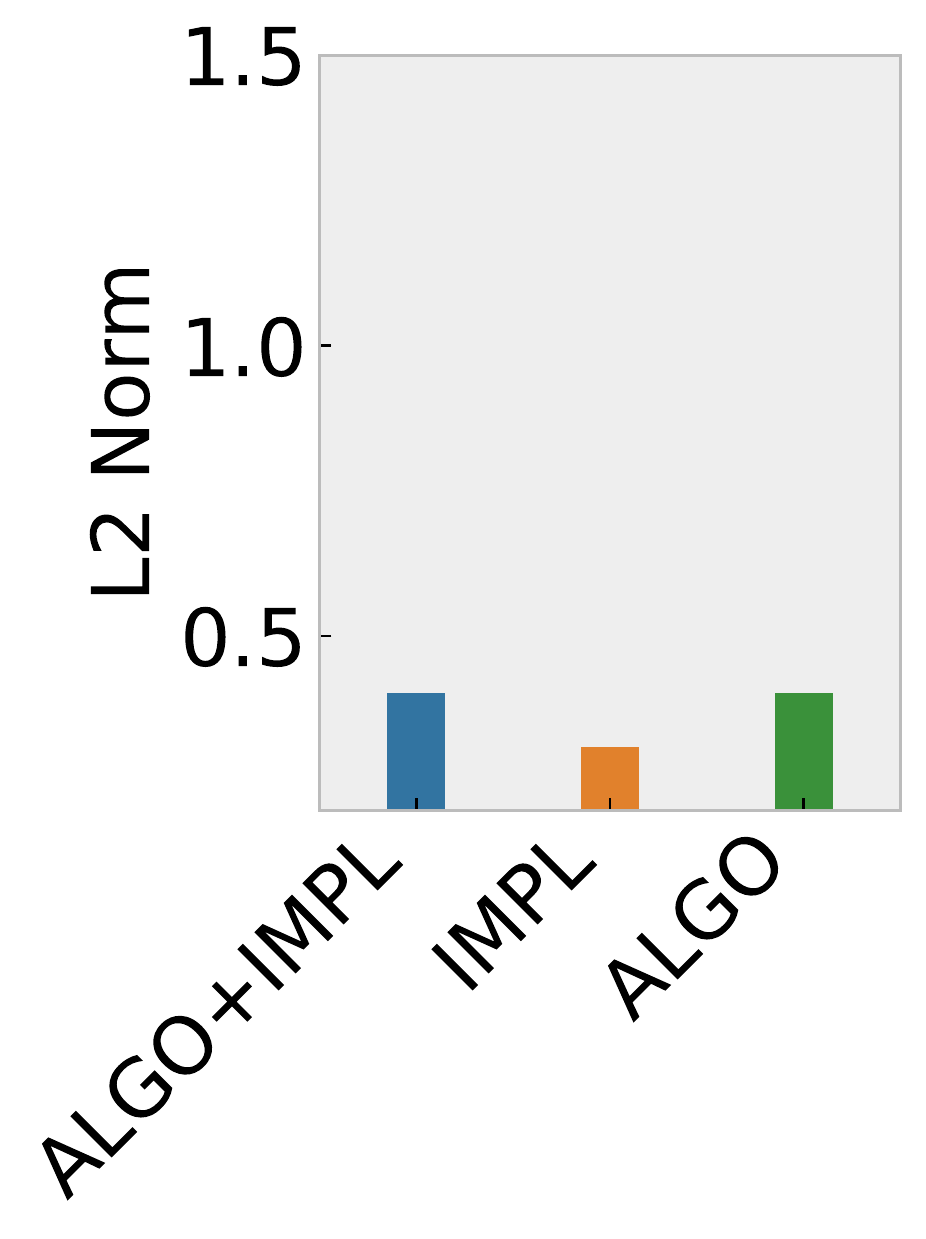}
        \end{subfigure}
        \caption{ResNet18 CIFAR-10}
    \end{subfigure}
    \\
    \begin{subfigure}{0.49\linewidth}
         \begin{subfigure}{0.32\linewidth}
    		\centering \includegraphics[width=0.99\linewidth]{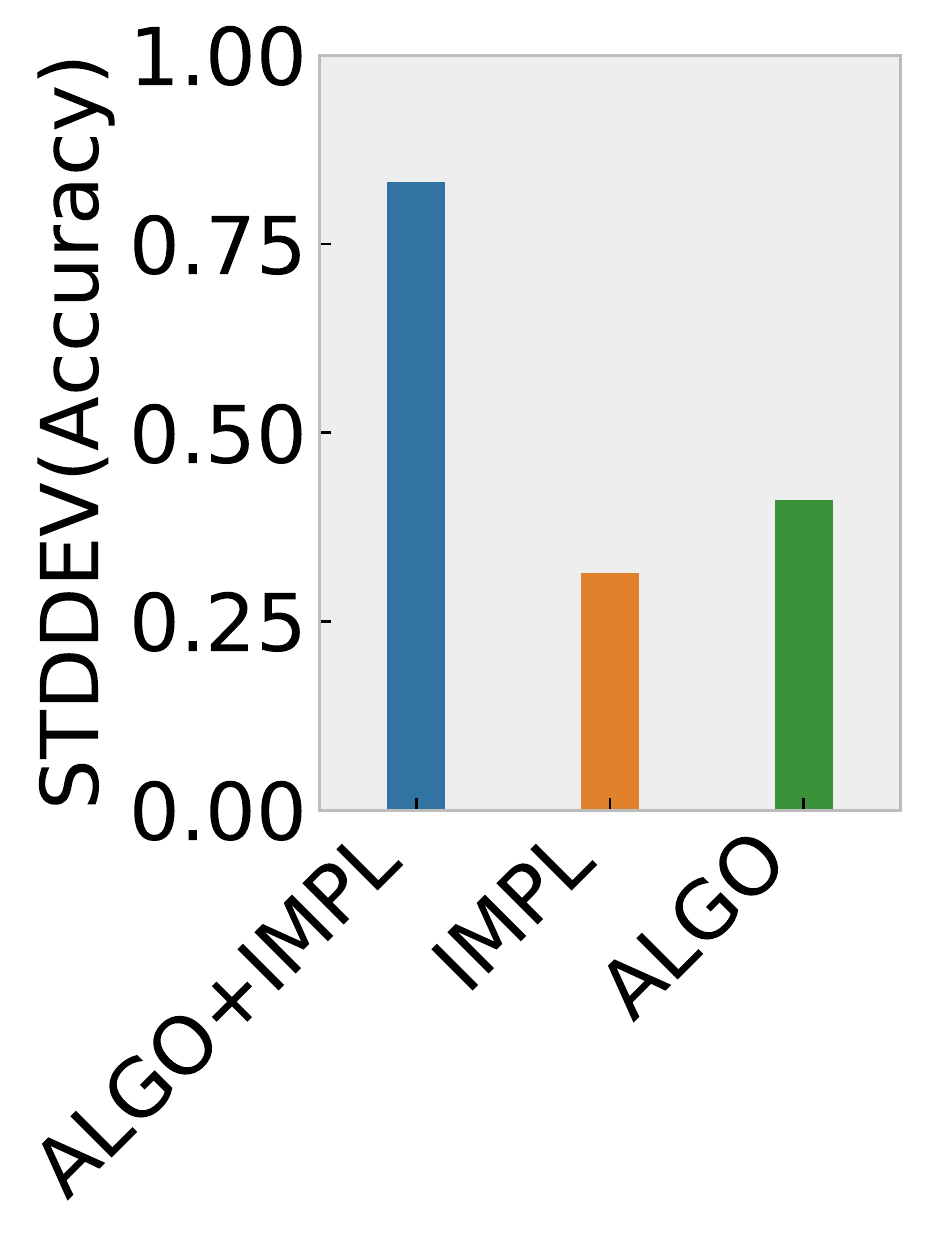}
        \end{subfigure}	
        \hspace{-0.5em}
        \begin{subfigure}{0.32\linewidth}
        	\centering	\includegraphics[width=0.99\linewidth]{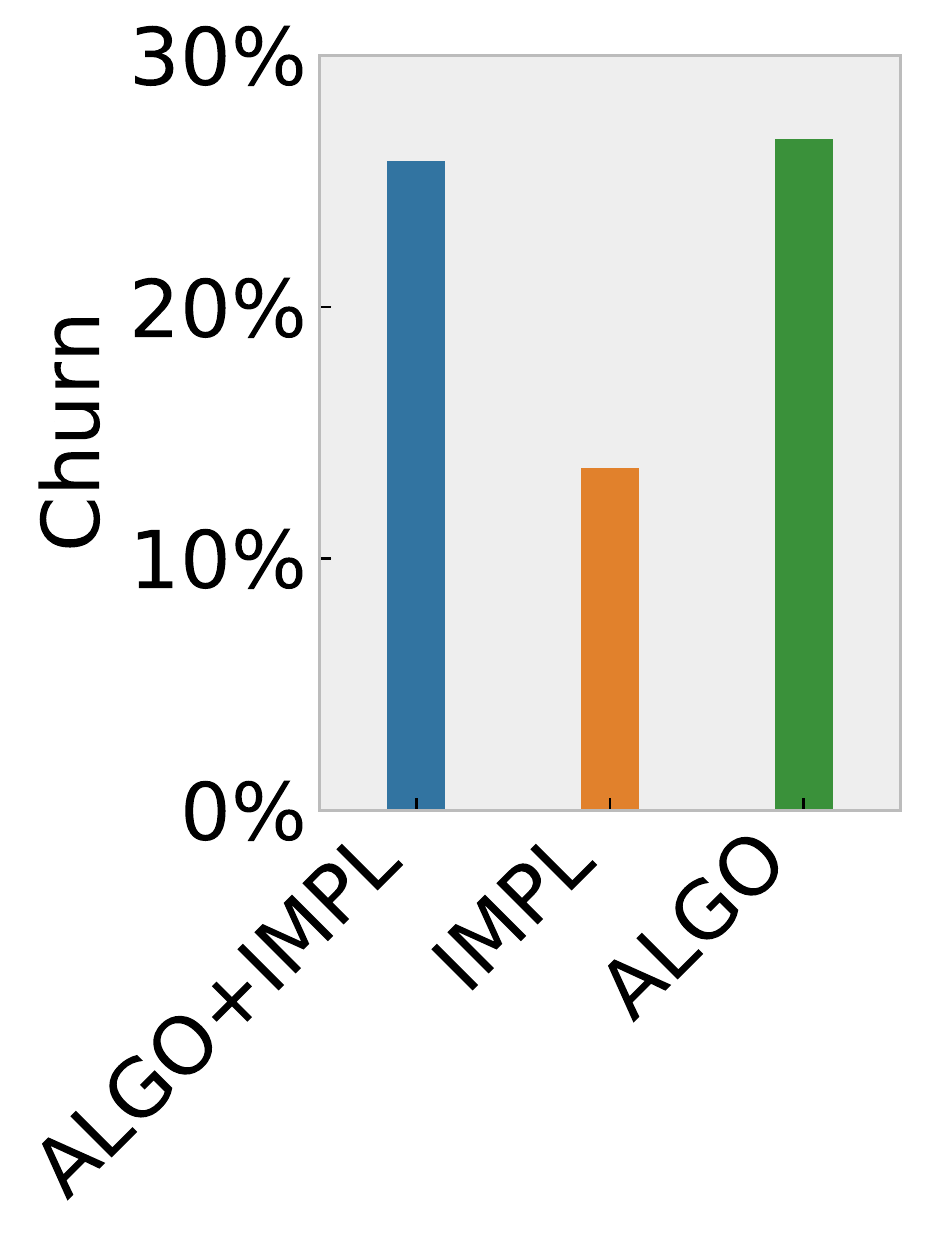}
        \end{subfigure}
        \hspace{-0.5em}
        \begin{subfigure}{0.32\linewidth}
        	\centering \includegraphics[width=0.99\linewidth]{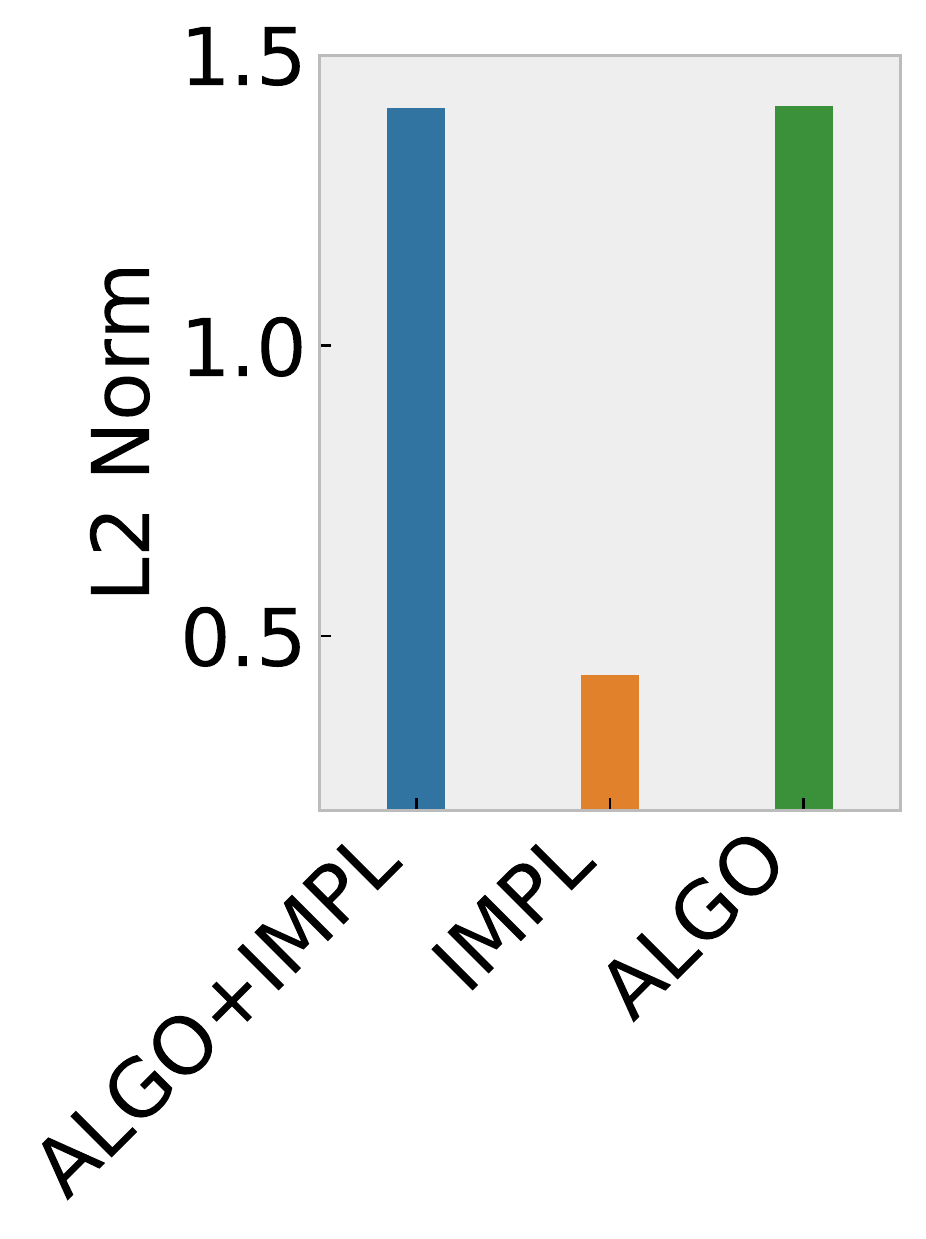}
        \end{subfigure}
        \caption{ResNet18 CIFAR-100}
    \end{subfigure}

	\end{sc}
	\end{small}
	\caption{
	   Comparison of impact of different source of noise across on four tasks trained on P100}
	\label{fig:types_noise_p100}
\end{figure*}

\begin{figure*}[ht!]
	\centering
	\vskip 0.15in
    \begin{small}
    \begin{sc}
    \begin{subfigure}{0.49\linewidth}
        \begin{subfigure}{0.32\linewidth}
		    \centering \includegraphics[width=0.99\linewidth]{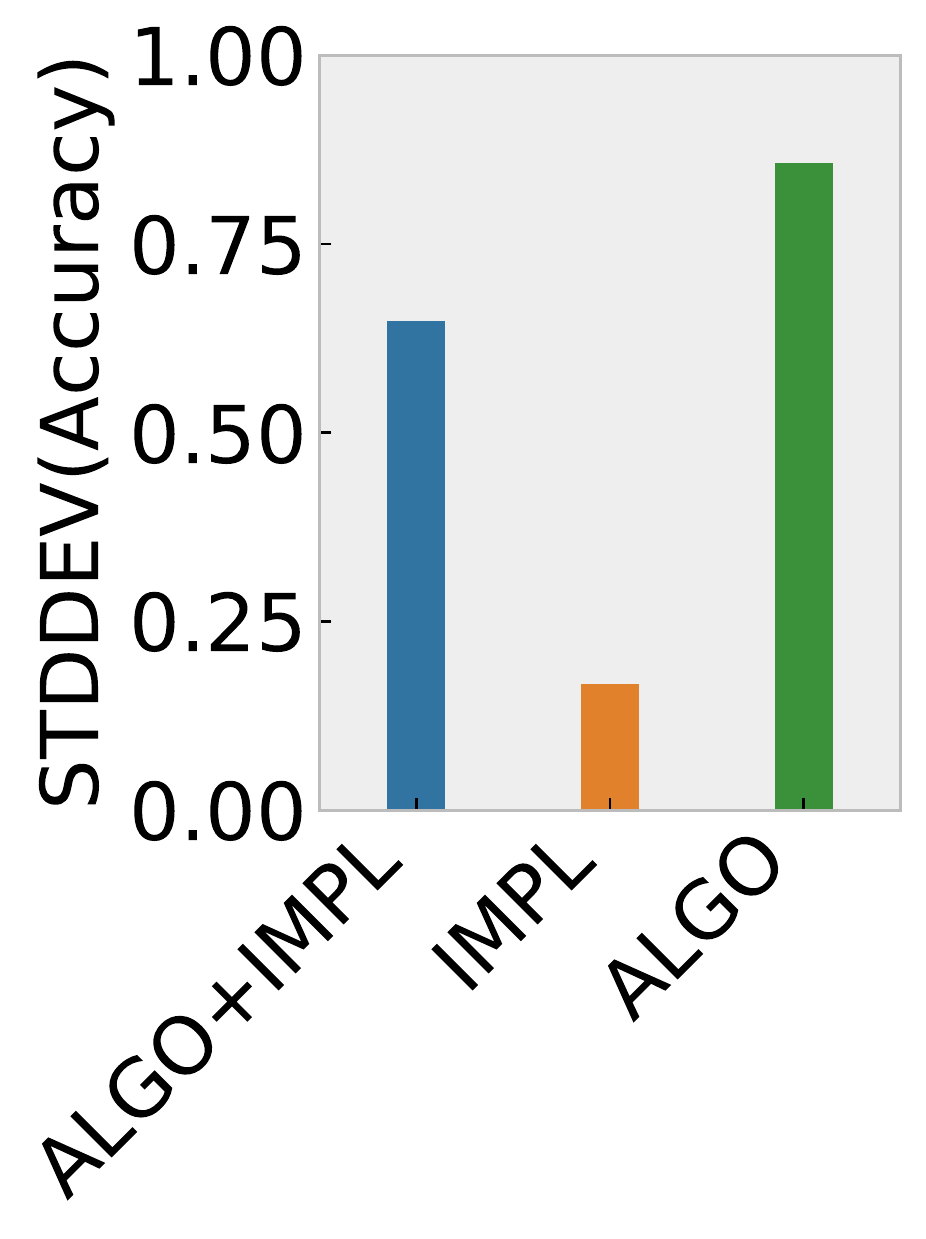}
        \end{subfigure}	
        \hspace{-0.5em}
        \begin{subfigure}{0.32\linewidth}
        	\centering \includegraphics[width=0.99\linewidth]{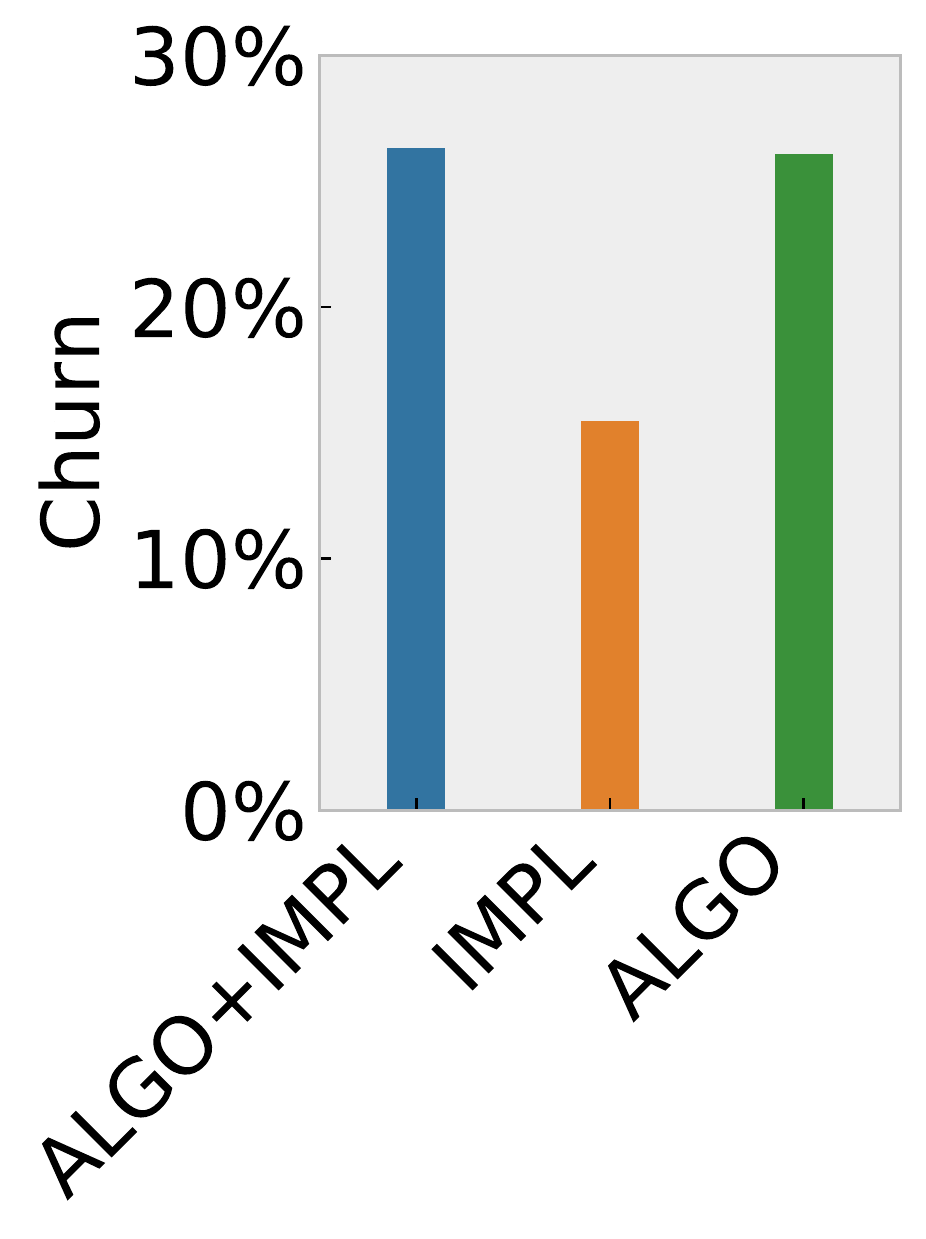}
        \end{subfigure}
        \hspace{-0.5em}
        \begin{subfigure}{0.32\linewidth}
        	\centering \includegraphics[width=0.99\linewidth]{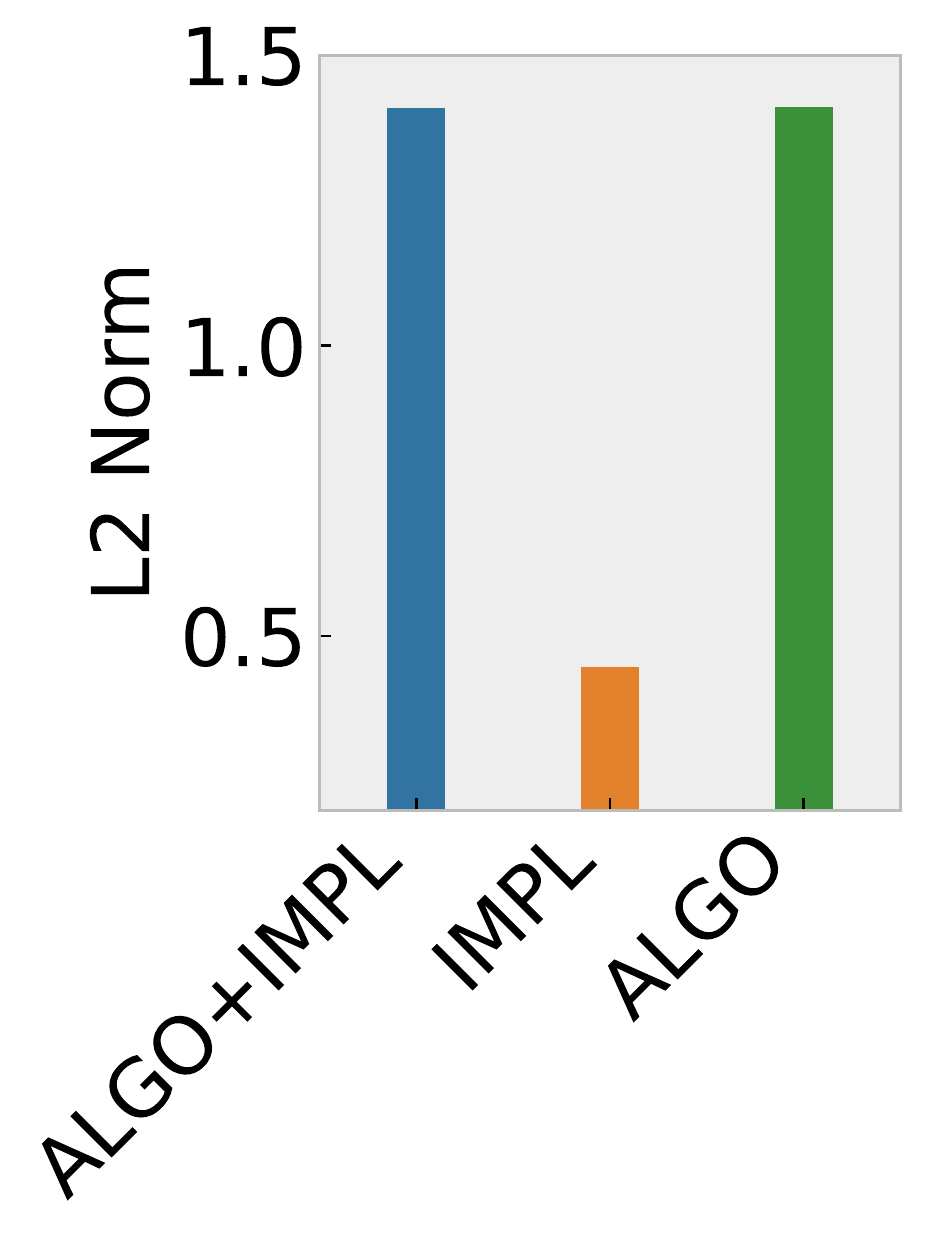}
        \end{subfigure}
        \caption{SmallCNN CIFAR-10}
    \end{subfigure}
    \\
    \begin{subfigure}{0.49\linewidth}
        \begin{subfigure}{0.32\linewidth}
    		\centering \includegraphics[width=0.99\linewidth]{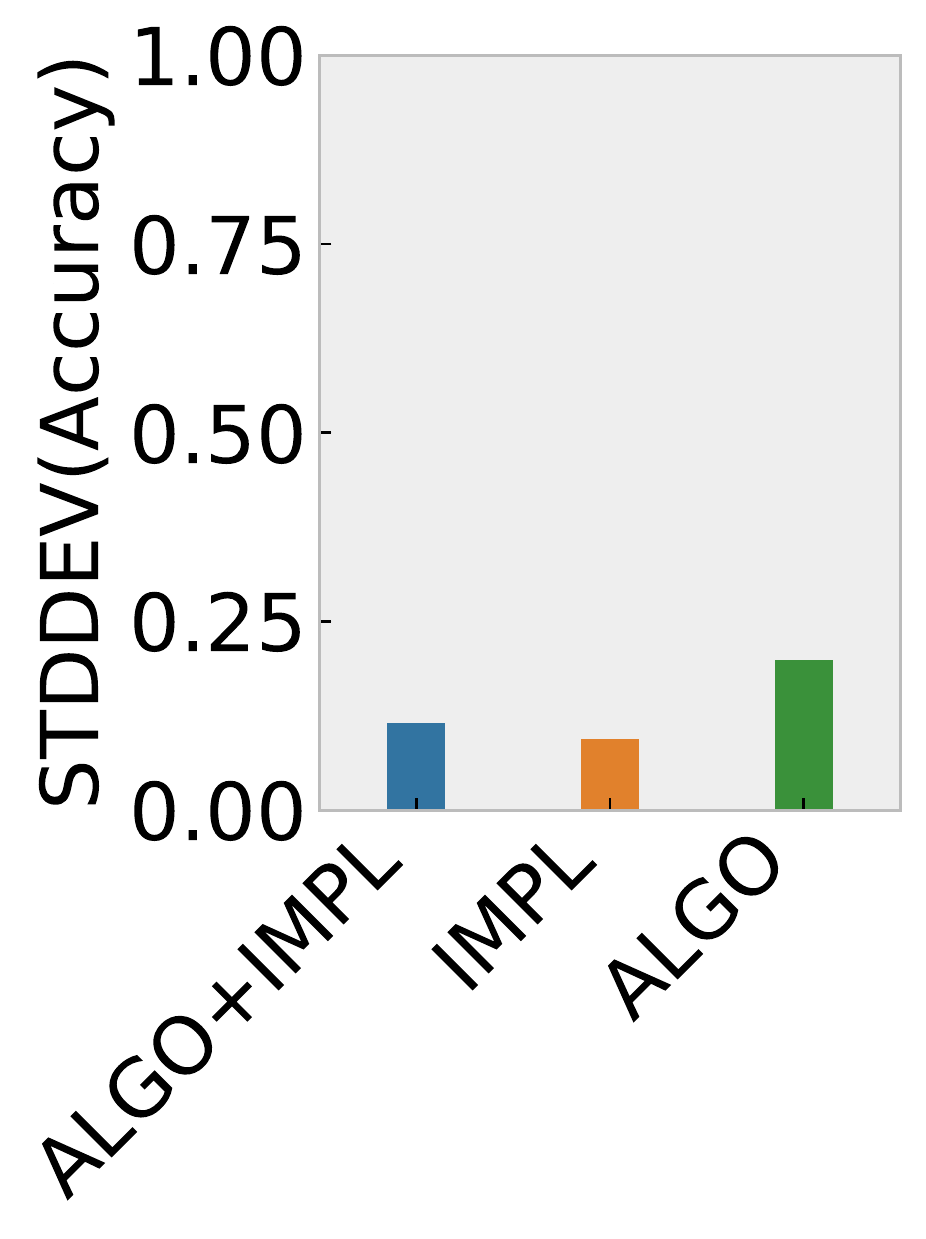}
        \end{subfigure}	
        \hspace{-0.5em}
        \begin{subfigure}{0.32\linewidth}
        	\centering	\includegraphics[width=0.99\linewidth]{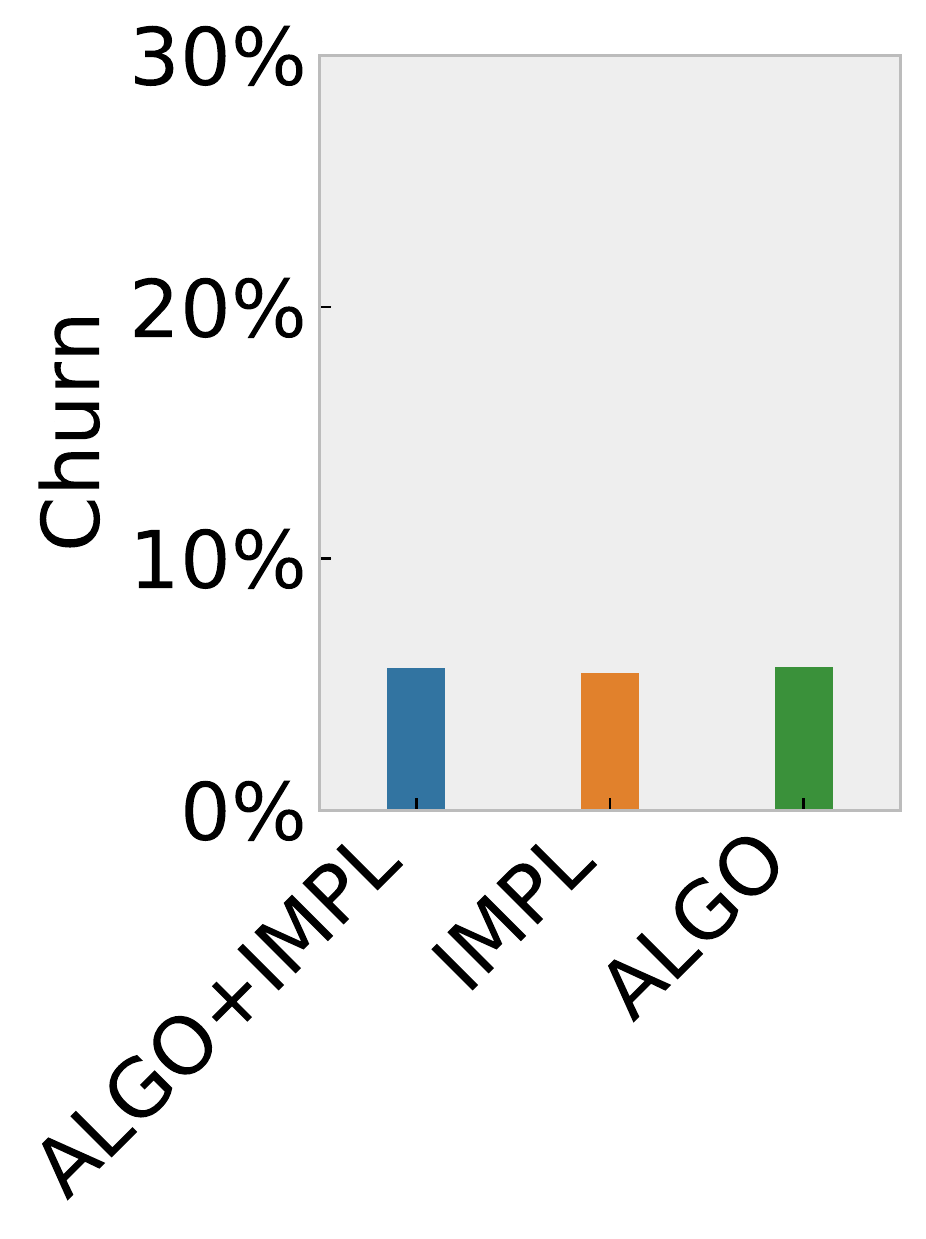}
        \end{subfigure}
        \hspace{-0.5em}
        \begin{subfigure}{0.32\linewidth}
        	\centering \includegraphics[width=0.99\linewidth]{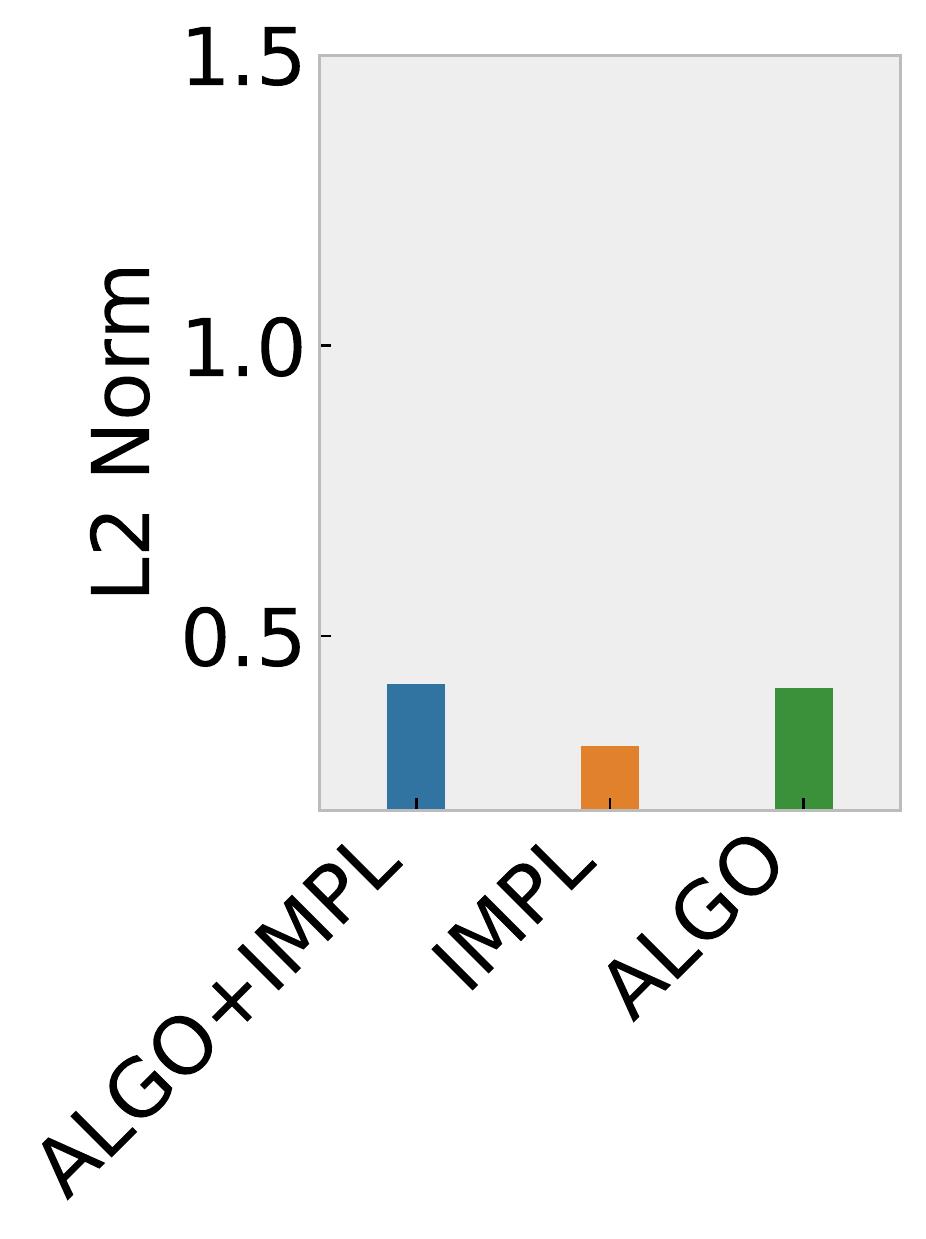}
        \end{subfigure}
        \caption{ResNet18 CIFAR-10}
    \end{subfigure}
    \\
    \begin{subfigure}{0.49\linewidth}
         \begin{subfigure}{0.32\linewidth}
    		\centering \includegraphics[width=0.99\linewidth]{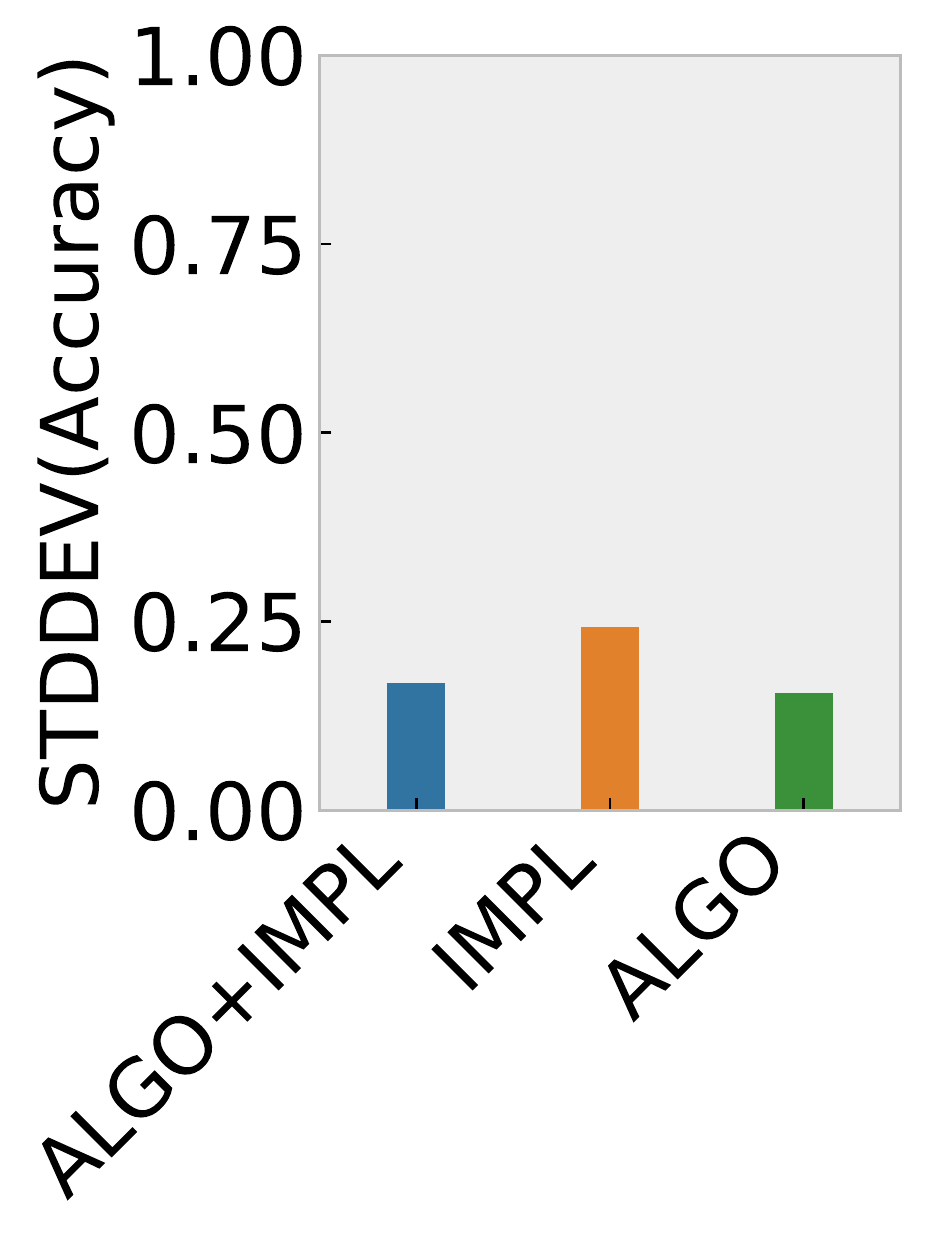}
        \end{subfigure}	
        \hspace{-0.5em}
        \begin{subfigure}{0.32\linewidth}
        	\centering	\includegraphics[width=0.99\linewidth]{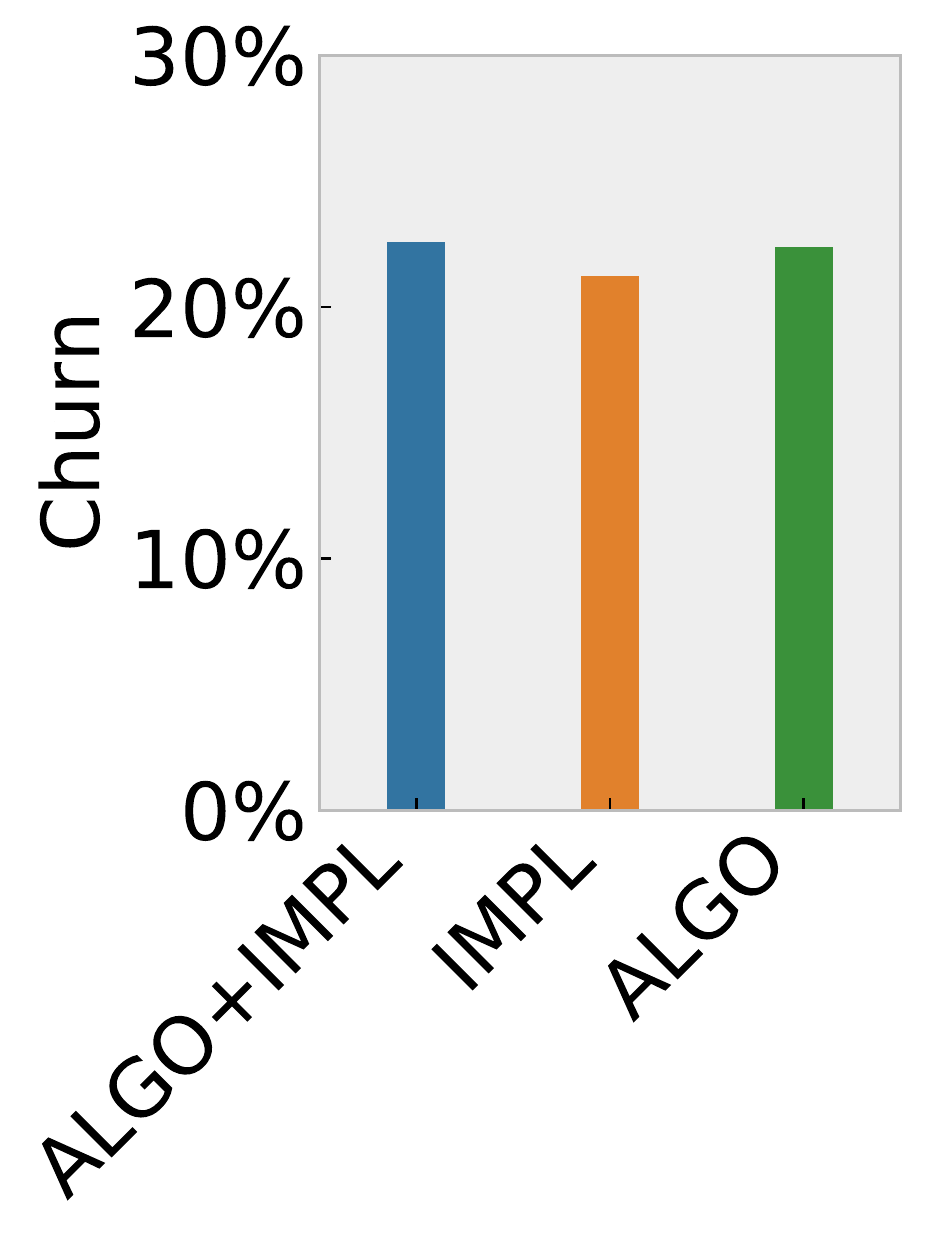}
        \end{subfigure}
        \hspace{-0.5em}
        \begin{subfigure}{0.32\linewidth}
        	\centering \includegraphics[width=0.99\linewidth]{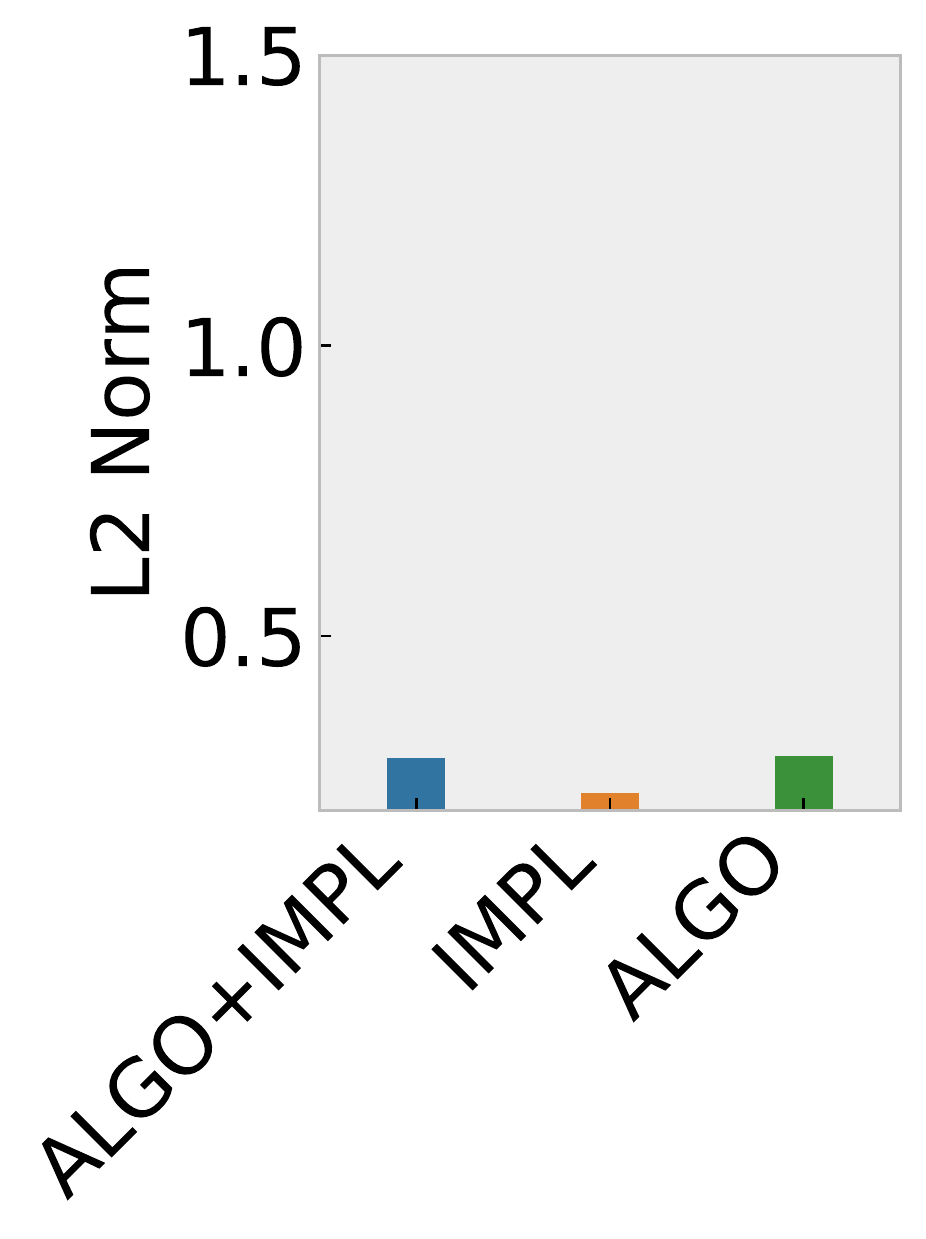}
        \end{subfigure}
        \caption{ResNet18 CIFAR-100}
    \end{subfigure}

	\end{sc}
	\end{small}
	\caption{
	   Comparison of impact of different source of noise across on four tasks trained on RTX5000}
	\label{fig:types_noise_rtx5000}
\end{figure*}

\textbf{Number of CUDA Cores} In Figure~\ref{fig:cifar100_noise_over_gpus}, we compare all hardware types we evaluate on CIFAR-100. In the appendix, we include additional breakdowns for each dataset/model/hardware evaluated (Figure~\ref{fig:types_noise_p100} and Figure~\ref{fig:types_noise_rtx5000}). For all GPUs we evaluate, V100 results in larger divergence under implementation noise in terms of both \texttt{churn} and \texttt{l2}. We attribute this to the relatively larger number of CUDA cores in V100 GPUs than either P100 and RTX5000, which suggests increased parallelism is a key driver of implementation noise. 

\textbf{Accelerator comparison} Surprisingly, we find that \texttt{IMPL} impact on \texttt{churn} and \texttt{l2} is still high for RTX5000 Tensor Cores which employ systolic arrays similar to TPUs to accelerate computation. The high level of \texttt{IMPL} despite the systolic design appears to be due to the reliance of Tensor Cores on non-deterministic CUDA cores on GPU for computations that not supported. Thus, model training leveraging Tensor Cores computation remains non-deterministic and is introducing a similar level of noise compared to CUDA Cores.


In Figure~\ref{fig:cifar100_noise_over_gpus}, it is visible that for \texttt{ALGO+IMPL} TPUs incurs a lower level of churn and l2 in weights compared to GPUs. This difference is due to the inherently deterministic design of TPUs, such that any stochasticity is only introduced algorithmic factors even under \texttt{ALGO+IMPL} setting. We oberve that while TPU lower \texttt{churn} and \texttt{l2} relative to GPUs, there is not a pronounced impact on \texttt{stdev}. This is consistently with our wider observation across experiments, we note that removing individual sources of noise tends to slightly reduce \texttt{churn} and \texttt{l2}, but does not have an observable relationship with \texttt{stddev} which appears far more sensitive to the presence of \emph{any} source of noise.

\textbf{Non-determinism based upon differences in ordering} 
Both GPUs and TPUs can introduce hardware noise because of it. In Figure~\ref{fig:ordering-nondeterminism}, we train ten small CNNs on CIFAR-10 dataset for each batch size, with all source of noise fixed except data shuffling order. When the batch is 50000, the full dataset is packed into a signal training batch, mathematically in this case all models should produce identical result. Interestingly, we still observe divergence of predictions between end runs for all batch size we evaluate. TPUs are designed for single-threaded, deterministic execution mode but are not ensured to be deterministic to ordering in data. This is because the difference in input data order will result in different float-point accumulation order in gradients accumulation stage thus introducing latent implementation noise.


\begin{figure*}[t]
	\vskip 0.15in
	\centering
\begin{small}
	\begin{sc}
    	\begin{subfigure}{0.45\linewidth}
    		\centering
        	\includegraphics[width=0.75\columnwidth]{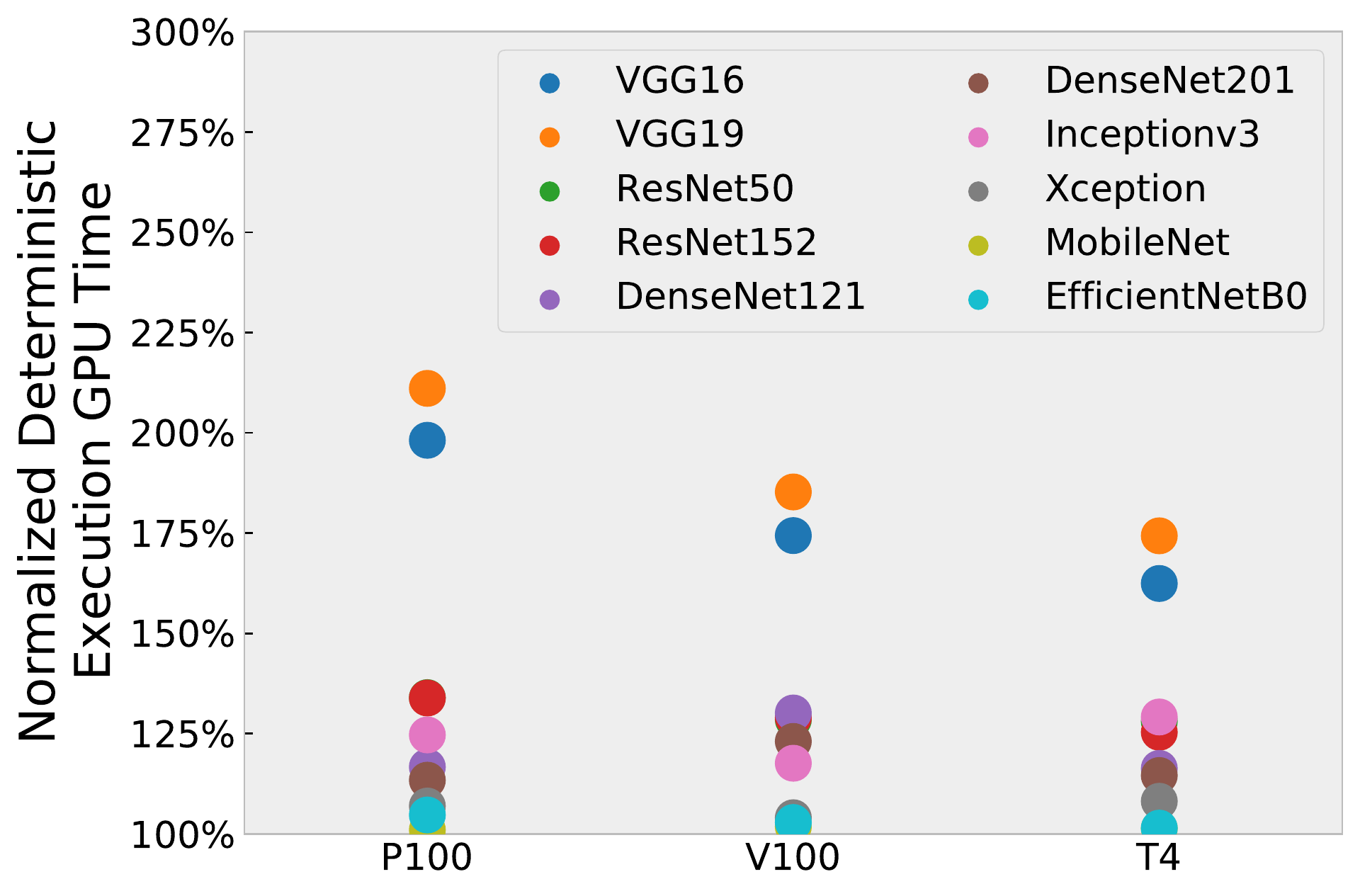}
    	    \caption{Across Networks}
            \label{fig:reducible_error}
    	\end{subfigure}
    	\begin{subfigure}{0.45\linewidth}
            \centering
            \includegraphics[width=0.75\columnwidth]{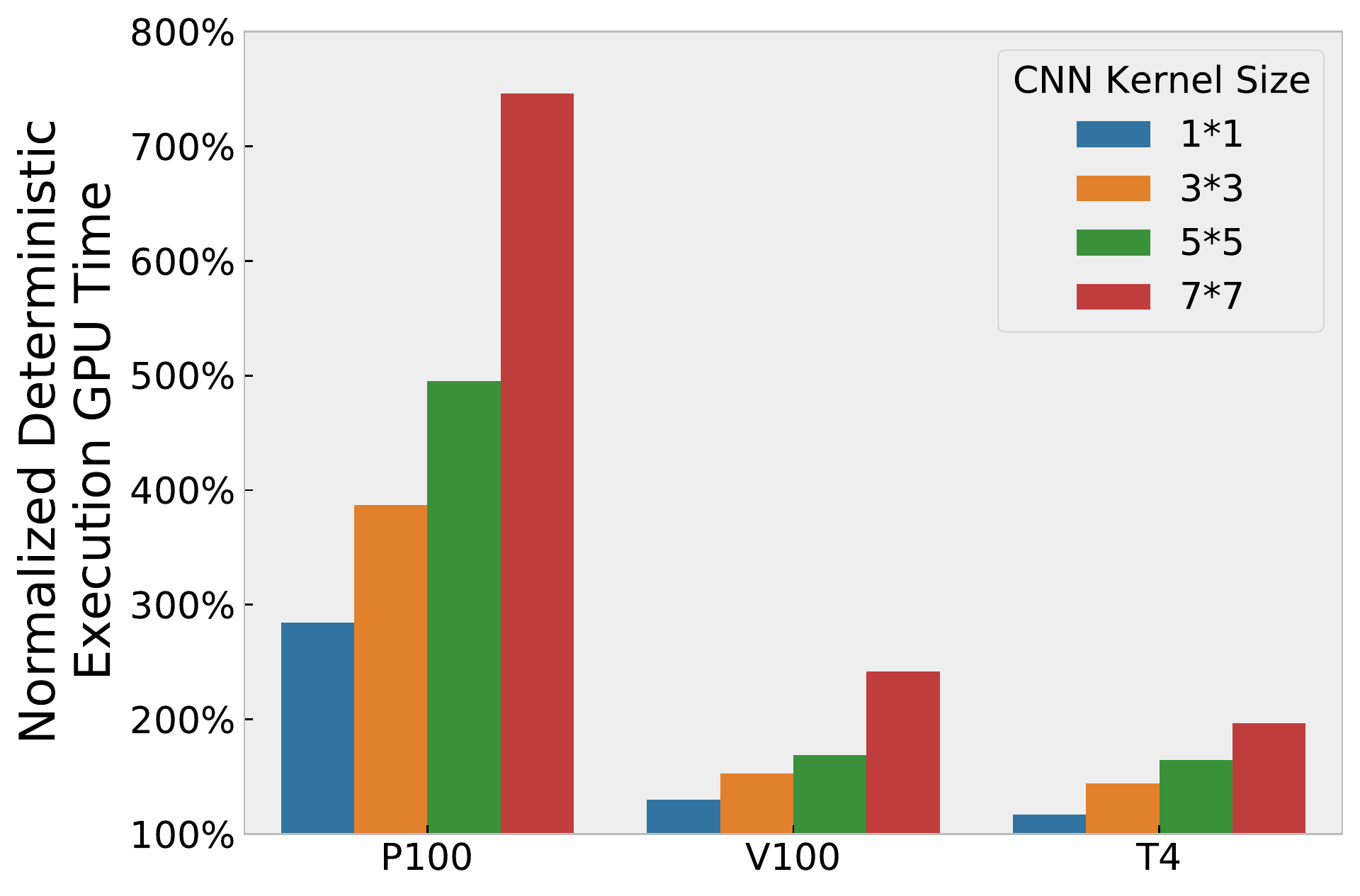}
            \caption{Across Kernels}
            \label{fig:irreduciblerror}
    	\end{subfigure}
	\end{sc}
\end{small}
	\caption{Comparison of GPU overhead of deterministic setting relative to non-deterministic training setting on \textbf{Left:} Ten widely used neural networks, \textbf{Right:} A six-layers medium CNN (Appendix~\ref{sec:appendix:smallcnn}) plugged with different size convolution kernels. }
	\label{fig:overhead_controlling_noise}
		\vskip -0.15in
\end{figure*}

\section{Results: The Cost of Ensuring Determinism}\label{sec:result:overhead}

Algorithmic noise can be controlled by simply setting a fixed random seed. In contrast, controlling implementation noise comes with non-negligible overhead. Most popular deep learning frameworks leverage cuDNN for high-performance computation on GPU. Some convolution algorithm implementation in cuDNN are designed to trade determinism for execution speed. Thus, the cost of controlling implementation noise should be thoroughly analyzed.

\textbf{Profiling Experiments} We profile the overhead of deterministic settings relative to normal training (\texttt{ALGO} + \texttt{IMPL}) by measuring GPU time spend on CUDA kernel computation using nvprof profiler \citep{nvprof}. This metric is well suited for our experiments since it is focusing how much time GPU spends on computation and excludes time intervals we do not care such as latency on data input pipeline. We select networks that are widely used such as MobileNet \citep{howard2017mobilenets}, EfficientNets \citep{tan2020efficientnet} , DenseNet-121/201 \citep{densenet}, VGG-16/19 \citep{Simonyan15} and ResNet-50/152 \citep{resnet}. We profile all models on ImageNet dataset with input shape 224*224 and batch size of 64. 

\textbf{How does the choice of model architecture impact overhead?}
Figure~\ref{fig:overhead_controlling_noise}~(a) shows the relative deterministic overhead of a variety of popular CNN models. VGG-19 has the most significant overhead among the models we profiled on all GPUs, with a $185\%$ relative GPU time compared to non-deterministic counterpart on V100 whereas MobileNet has only $101\%$ relative GPU time compared to to non-deterministic counterpart. P100 and T4 also present a large variation of deterministic overhead associate with different model architectures with range $101\%\sim211\%$ and $101\%\sim196\%$ respectively.

\textbf{The role of filter size}  To further understand the relative overhead of variation in size of convolutional filters, we evaluate across different kernel sizes using a six layer medium CNN (Appendix~\ref{sec:appendix:smallcnn}). Assembled with convolution kernel size ranging from $1*1$ to $7*7$. As show in Figure~\ref{fig:overhead_controlling_noise}~(b), the GPU overhead time is remarkably sensitive to the size of kernel, with $284\%\sim746\%$ on P100, $129\%\sim241\%$ on V100, and $117\%\sim196\%$ on T4 respectively. For all kernel size we evaluate on each GPU, larger kernel size is always comes with larger overhead.

\textbf{How does the choice of hardware impact overhead?}
We observe that GPU architecture overhead varies considerably and is highly dependent on model design choices. For example, as shown in Figure~\ref{fig:overhead_controlling_noise}~(b), we observe overhead for a 7*7 kernel relative to default mode is up to 746\%, 241\%, and 196\% on \textit{P100}, \textit{V100}, and \textit{T4} respectively. Consistently, across all models ranging from six-layer medium CNN to ResNet50, GPUs with older \textit{Pascal} architecture (P100) evidence higher overhead than GPUs with later \textit{Volta} (V100) and \textit{Turing} (T4) architecture.  

The large and highly variable overhead suggests deterministic training comes with non-negligible overhead on which researchers should weigh the price and benefit based on their tolerance to uncertainty. However, even the minimum observed overhead poses significant hurdles to efficient training. In Figure~\ref{fig:overhead_overview}, we plot the time spent using the Top-20 kernels selected across 100 steps of training. The more skewed time allocation of deterministic mode shows the heavy dependency on a narrower set of kernels instead tuning the best one heuristically.  This cost can be attributed to the narrow range of kernels the compiler is forced to use when deterministic training is selected.

\section{Related Work} \label{sec:related_work}

\textbf{Reproducibility in machine learning} As numerous works have pointed out \citep{Goodman341ps12,Gundersen2018StateOT,barba2018terminologies,risnotr}, the word reproducibility can correspond to very different standards, ranging from the ability to reproduce statistically similar values \citep{raff2019step,McDermott2019ReproducibilityIM,7840958}, to successfully executing code \citep{Collberg2016}, to the ability to reproduce a relative relationship (a model remains state of art even when the experimental set-up is changed) \citep{pmlr-v97-bouthillier19a}. In this work, we are concerned with replicability, a subset of reproducibility where the standard is reproducing the exact results given the same experimental framework. Advances in tooling have aimed to simplify replication, ranging from shareable notebooks \citep{soton403913}, dockerization \citep{10.5555/2600239.2600241}, machine learning platforms where code and data is easily shareable \citep{9041744} and software patches to ensure determinism for a subset of operations. Less mature ideas include research around automatic generation of code from research papers \citep{sethi2017dlpaper2code}. In the computer architecture research community, researchers have proposed several deterministic GPU architectures \citep{gpudet,dab} to boost the reproducibility and debuggability of GPU workloads. 

\textbf{Impact of Algorithmic Factors} A substantial amount of work has considered the impact of different sources of randomness introduced by algorithm design choices. Several works have evaluated the impact of a random seed, with \citep{impactofnondeterminismrl} evaluating the role of random initialization in reinforcement learning, and \citep{madhyastha-jain-2019-model} measuring how random seeds impact explanations for NLP tasks provided by interpretability methods. \citep{summers2021nondeterminism} benchmark the separate impact of choices of initialization, data shuffling and augmentation. Work mentioned thus far is focused on how design choices that introduce randomness impact training. However, there is a wider body of scholarship that has focused on sensitivity to non-stochastic factors including choice of activation function and depth of model \citep{Snapp2021,shamir2020smooth},  hyper-parameter choices \citep{lucic2018gans,Henderson2017,kadlec2017}, the use of data parallelism \citep{shallue2019measuring} and test set construction \citep{sogaard-etal-2021-need,lazaridou2021pitfalls,Melis2018OnTS}.

\textbf{Impact of Software Dependencies}  \citep{Hong2013} evaluate the role of different compilers for the specialized task of weather simulation. Recent work by \citep{Pham2020} and
\citep{9266043} in the machine learning domain evaluates the impact of randomness introduced by popular deep neural network libraries (Pytorch, CNTK, Theano and Tensorflow). \citep{9266043} evaluates a segmentation task for mouse neo-cortex data and MNIST on LeNet \citep{726791}. \citep{Pham2020} finds the biggest variance across all deep learning libraries on LeNet5. These works and others only evaluate the role of software dependencies on a single type of hardware. Our contribution is the first to our knowledge to vary the hardware, and measure the cost of ensuring determinism across different types of hardware. 

\textbf{Trade-off with fairness objectives} Recent work \citep{HOOKER2021100241,yona2021whos,damour2020underspecification,2019arXiv191105248H} has identified that models with similar top-line metrics can evidence unacceptable performance on subsets of the distribution. Design choices such as compression \citep{hooker2020characterising} and privacy \citep{Cummings2019} can impact disparate impact on sensitive attributes. However, ours is the first to our knowledge that evaluates the impact of tooling and sources of randomness on disparate harm. 

\section{Discussion and Future Work} \label{sec:future_work}
There has been increasing urgency to ensure non-determinism in deep neural network training. However, in the rush to mitigating or eliminating noise in deep learning system, a natural question that have not been discussed thoroughly would be: \textit{What is the impact of noise? Does the impact merit the cost of controlling it?}.

\textbf{Limitations} In this work, our focus is evaluating the impact of tooling in a non-distributed setting. However, increasingly training deep neural networks involves data and model parallelism \citep{meshtensrflow, Langer_2020}, partition over optimizer state \citep{zero}, and asynchronous gradients update \citep{parameterserver}. An important area of future work involves understanding how distributed training impacts model stability. 


\section{Conclusion}
In this work, we seek to characterize the impact and cost of controlling noise at \emph{all} levels of the technical stack. We empirically demonstrate that both algorithmic and implementation noise are significant sources of noise. Thus, simply removing noise from one part of the technical stack is not a robust way to improve training stability. Secondly, we show that even with minimal changes to top-line metrics, there is a disproportionately impact on sub-group performance which can incur fairness trade-offs when protected attributes are underrepresented. Finally, we evaluate the cost of ensuring determinism and find it is highly variable and dependent on hardware type and model design choices.

 

\bibliographystyle{abbrvnat}
\setcitestyle{authoryear,open={((},close={))}} 
\bibliography{reference}

\clearpage
\renewcommand\thesubsection{\Alph{subsection}}
\section{Appendix} \label{sec:appendix}

\subsection{Sources of Randomness During Deep Neural Network Training} \label{subsec:appendix_sources_randomness}

\textbf{Algorithmic Factors}~(\texttt{ALGO}) includes model design choices which are stochastic by design. Often, there are widely used implementation choices as introducing stochasticity to deep neural network training has been found to improve top-line metrics:
\begin{itemize}
\itemsep0em
\item \textbf{Random Initialization}~-~the weights of a deep neural network are randomly initialized, typically with the goal is maintaining variance of activations within a narrow range at the beginning of training to avoid gradient saturation \citep{pmlr-v9-glorot10a,resnet}.
\item \textbf{Data augmentation}~-~the quality of a trained model depends upon the training data. Often, when faced with limited data an effective strategy is to generate new samples by applying stochastic transformations to the input data \citep{kukacka2017regularization,Hern_ndez_Garc_a_2018}. Examples of stochastic data augmentation include random crops, noise injection, and random distortions to color channels \citep{dwibedi2017cut,zhong2017random}. 
\item \textbf{Data shuffling and ordering}~-~for mini-batch stochastic gradient optimization, datasets are typically shuffled randomly during training and batched into a subset of observations.. Thus, each training process will observe a different ordering of inputs. Batching examples introduces noise through stochastic mini-batch gradient descent \citep{smith2018dont}. Even when batching is not used (all data is processed in a single batch), a difference in ordering can introduce stochasticity that may introduce security vulnerabilities \citep{shumailov2021manipulating}.
\item \textbf{Stochastic Layers}~-~techniques such as dropout which entails randomly dropping a subset of weights each iteration \citep{Srivastava2014,hinton2012improving,pmlr-v28-wan13}, noisy activation functions \citep{10.5555/3104322.3104425} or variable length backpropagration through time \citep{merity2017regularizing}.
\end{itemize}



\subsection{Training Methodology}\label{sec:appendix:training_methodology}

We employ random crop and flip for data augmentation on all experiments except experiments on CelebA dataset. 

\textbf{CIFAR-10 and CIFAR-100} \citep{Krizhevsky09learningmultiple} We train a small CNN on CIFAR-10 which consists of three convolutional layers, followed by a dense layer and a output layer. Additionally, we evaluate both CIFAR-10 and CIFAR-100 on ResNet-18. For all networks, we train for 200 epochs with a batch size of 128 and $4e-4$ learning rate which decays by a factor of ten every 50 epochs.

\textbf{CelebA} \citep{celeba} CelebA dataset consist of $\sim$200K celebrity's facial images, each image associated with labels with forty binary attributes such as identifying hair color, gender, age.  Our goal is to understand the implications of noise on model bias and fairness considerations. Thus, we focus attention on two protected unitary attributes \texttt{Male}, \texttt{Female} and \texttt{Young} and \texttt{Old}. Our goal is to understand the implications of noise on model bias and fairness considerations. we measure standard deviation of sub-group accuracy, false positive rate (FPR) and false negative rate (FNR). We train ResNet18 on CelebA dataset for 20 epochs with batch size of 128 and learning rate of $1e-3$ decays by a factor of ten every 5 epochs.

\textbf{ImageNet} \citep{IRSVRC} On ImageNet dataset, we train ResNet-50 for 90 epochs with batch size of 256 with learning rate $0.1$ using SGD optimizer with momentum of 0.9, the learning rate is warming up in the first epoch and using cosine decay in the following epochs. We conduct out experiment on Imagenet dataset based on ResNet50 implementation from Tensorflow Model Garden \footnote{https://github.com/tensorflow/models}

\clearpage
\subsection{CNN Architecture}\label{sec:appendix:smallcnn}
Architecture of three-layer small CNN and six-layer medium CNN. Downsampling is performed in pooling layers, all convolutional layers are using stride=1. For six-layer small CNN, kernel size $X$ can be 1, 3, 5, and 7.

\input{tables/appendix_smallcnn}

\clearpage
\subsection{Comparison of Impact of Different Source of Noise} \label{sec:appendix:compounding}

\begin{table*}[t]
\begin{center}
\begin{small}
\begin{tabular}{c|c|ccc}
\toprule
\textbf{Dataset} & \textbf{Training/Test Split} & \textbf{Number Classes} \\
Cifar-10 & 50000/10000 & 10 \\
Cifar-100 & 50000/10000 & 100 \\
ImageNet & 1281167/50000 & 1000 \\
CelebA & 162770/19962 & 40 (Multi-label) \\
\bottomrule
\end{tabular}
\end{small}
 \end{center}
 \caption{Overview of each dataset benchmarked.}
 \label{table:dataset_bias} \end{table*}


 \begin{table*}
    \centering
      \begin{tabular}{cccc}
        \toprule
    \textbf{Subgroup}  & \texttt{ALGO+IMPL} & \texttt{ALGO} & \texttt{IMPL} \\
     \midrule
        \midrule
         & \multicolumn{3}{c}{STDDEV(Accuracy)}  \\
        All	 & 0.045 (1X)  & 0.051 (1X) & 0.090 (1X)      \\  
        MALE & 0.049	(1.07X) & 0.048 (0.94X) &  0.058 (0.64X) \\
        FEMALE	    & 0.062	(1.36X) & 0.083 (1.62X) &  0.126 (1.39X)  \\
        YOUNG	     & 0.050 (1.10X) & 0.047 (0.93X)&  0.091 (1.00X)  \\
        OLD	        & 0.151	(\textbf{3.31X}) & 0.094 (1.83X) & 0.214 (\textbf{2.36X})    \\
        \midrule
        \midrule
       & \multicolumn{3}{c}{STDDEV(FPR)} \\
      \midrule All	 & 0.077 (1X) & 0.051 (1X) & 0.070 (1X)    \\  
        MALE	  & 0.039 (0.50X) & 0.052 (1.01X) & 0.043 (0.61X)           \\
        FEMALE	 & 0.133 (1.71X)  & 0.094 (1.81X) & 0.103 (1.48X)          \\
        YOUNG	 & 0.077 (1.00X)  & 0.051 (0.99X) & 0.065 (0.93X)          \\
        OLD	     & 0.122 (1.57X) & 0.093 (1.81X) & 0.155  (\textbf{2.21X})          \\
        \midrule
              \midrule
              &   \multicolumn{3}{c}{STDDEV(FNR)} \\
          \midrule
      All	& 0.537 (1X) & 0.389 (1X) &  0.445 (1X) \\  
        MALE &	  2.475 (\textbf{4.60X}) & 1.816 (\textbf{4.66X}) & 1.610 (\textbf{3.61X}) \\
        FEMALE	   & 0.527 (0.98X) & 0.349 (0.89X) & 0.399 (0.89X) \\
        YOUNG	  &  0.585 (1.08X) & 0.430 (1.10X) & 0.566 (1.27X) \\
        OLD	        &  0.815 (1.51X)  & 0.335 (0.86X) & 0.939 (\textbf{2.10X}) \\
        \bottomrule
    \end{tabular}
        \caption{Standard deviation of mean accuracy, false positive rate (FPR), and false negative rate (FNR) across 10 models trained under baseline setting on the CelebA dataset (trained on V100 (using cuda cores)). Metrics are dis-aggregated across two binary dimensions Male/Female and Young/Old. In parentheses, we report relative scale of standard deviation metrics relative to overall dataset.}
    \label{appendix_table:ceblea_breakdown}
\end{table*}

\end{document}

%% file: tables/appendix_smallcnn.tex
\begin{table}[h]
  \vspace{3pt}
  \label{table:appendix:smallcnn}
  \centering
  \begin{tabular}{cc|cc}
    \toprule
    \multicolumn{2}{c}{Three-layer small CNN} & \multicolumn{2}{c}{Six-layer medium CNN} \\
    Layer & Output Shape & Layer & Output Shape \\
    \midrule
    Input & $32*32*3$ & Input & $224*224*3$ \\
    \hline
    \\
    $\begin{bmatrix} Conv\:3*3 \\ Relu \\ MaxPool \end{bmatrix}$ & $\begin{matrix}16*16*16\end{matrix}$ & $\begin{bmatrix} Conv\:X*X \\ BN \\ Relu \\ MaxPool \end{bmatrix}$ & $\begin{matrix}112*112*16\end{matrix}$  \\
    \\

    $\begin{bmatrix} Conv\:3*3 \\ Relu \\ MaxPool \end{bmatrix}$ & $\begin{matrix}8*8*32\end{matrix}$ & $\begin{bmatrix} Conv\:X*X \\ BN \\ Relu \\ MaxPool \end{bmatrix}$ & $\begin{matrix}56*56*32\end{matrix}$  \\
    \\
    
    $\begin{bmatrix} Conv\:3*3 \\ Relu \\ MaxPool \end{bmatrix}$ & $\begin{matrix}4*4*32\end{matrix}$ & $\begin{bmatrix} Conv\:X*X \\ BN \\ Relu \\ MaxPool \end{bmatrix}$ & $\begin{matrix}28*28*64\end{matrix}$  \\
    \\
    
    &&$\begin{bmatrix} Conv\:X*X \\ BN \\ Relu \\ MaxPool \end{bmatrix}$ & $\begin{matrix}14*14*128\end{matrix}$  \\
    \\
    
    &&$\begin{bmatrix} Conv\:X*X \\ BN \\ Relu \\ MaxPool \end{bmatrix}$ & $\begin{matrix}7*7*256\end{matrix}$  \\
    \\
    
    &&$\begin{bmatrix} Conv\:X*X \\ BN \\ Relu \\ MaxPool \end{bmatrix}$ & $\begin{matrix}3*3*512\end{matrix}$  \\
    \\
    \hline
    \multicolumn{4}{c}{GlobalAveragePooling} \\
    \hline
    Dense & $\begin{matrix}32\end{matrix}$ &  &   \\
    Dense & $\begin{matrix}10\end{matrix}$ & Dense & $\begin{matrix}1000\end{matrix}$ \\
    \bottomrule
  \end{tabular}
\end{table}